\documentclass{article}

\usepackage{arxiv}

\usepackage[utf8]{inputenc} 
\usepackage[T1]{fontenc}    
\usepackage{hyperref}       
\usepackage{url}            
\usepackage{booktabs}       
\usepackage{amsfonts}       
\usepackage{nicefrac}       
\usepackage{microtype}      
\usepackage{lipsum}		
\usepackage{graphicx}
\usepackage[round]{natbib}
\usepackage{doi}

\usepackage{listings}

\usepackage{amsfonts}       
\usepackage{microtype}      
\usepackage{subcaption}
\usepackage{amsmath}
\usepackage[ruled,vlined,linesnumbered]{algorithm2e}
\usepackage{cleveref}
\DeclareRobustCommand{\abbrevcrefs}{%
\Crefname{appendix}{App.}{Apps.}%
\Crefname{section}{Sec.}{Secs.}%
\Crefname{equation}{Eq.}{Eqs.}%
\Crefname{figure}{Fig.}{Figs.}%
\Crefname{algorithm}{Alg.}{Algs.}%
\Crefname{tabular}{Tab.}{Tabs.}%
\Crefname{lemma}{Lem.}{Lems.}%
\Crefname{corollary}{Cor.}{Cors.}%
\Crefname{theorem}{Thm.}{Thms.}%
\Crefname{proposition}{Prop.}{Props.}%
\Crefname{line}{L.}{Ls.}%
\crefname{appendix}{app.}{apps.}%
\crefname{section}{sec.}{secs.}%
\crefname{equation}{eq.}{eqs.}%
\crefname{figure}{fig.}{figs.}%
\crefname{algorithm}{alg.}{algs.}%
\crefname{tabular}{tab.}{tabs.}%
\crefname{lemma}{lem.}{lems.}%
\crefname{corollary}{cor.}{cors.}%
\crefname{theorem}{thm.}{thms.}%
\crefname{proposition}{prop.}{props.}%
\crefname{line}{l.}{ls.}%
}

\DeclareRobustCommand{\Cshref}[1]{{\abbrevcrefs\Cref{#1}}}

\DeclareMathOperator*{\argmax}{arg\,max}

\def\cS{{\cal S}}
\def\cA{{\cal A}}

\def\cZ{{\cal Z}}

\def\eg{{\em e.g.}\xspace}

\def\fsc{\mathit{fsc}}

\mathchardef\mhyphen="2D

\usepackage{xcolor}
\usepackage[inline]{enumitem}
\usepackage{booktabs}       
\usepackage{multirow} 
\usepackage{array}   
\usepackage{siunitx} 
\usepackage{makecell}

\usepackage[most]{tcolorbox}

\newtcbtheorem[auto counter]{definitionbox}{Definition}{
    lower separated=false,
    colback=white!80!gray,
    colframe=white,
    fonttitle=\bfseries,
    colbacktitle=white!50!gray,
    coltitle=black,
    enhanced,
    boxed title style={colframe=black},
    attach boxed title to top left={xshift=0.5cm,yshift=-2mm},
}{def}

\title{Neural Value Iteration}

\author{ {Yang You} \\
    RACE\\
	UKAEA\\
	\And
    {Ufuk Çakır} \\
	Oxford Robotics Institute\\ University of Oxford\\
	\And
    {Alex Schutz} \\
	Oxford Robotics Institute\\ University of Oxford\\
    \And
    {Nick Hawes} \\
	Oxford Robotics Institute\\ University of Oxford\\
}

\date{}

\renewcommand{\shorttitle}{\textit{Neural Value Iteration}}

\begin{document}
\maketitle

\begin{abstract}
The value function of a POMDP exhibits the piecewise-linear-convex (PWLC) property and can be represented as a finite set of hyperplanes, known as $\alpha$-vectors.  
Most state-of-the-art POMDP solvers (offline planners) follow the point-based value iteration scheme, which performs Bellman backups on $\alpha$-vectors at reachable belief points until convergence.  
However, since each $\alpha$-vector is $|S|$-dimensional, these methods quickly become intractable for large-scale problems due to the prohibitive computational cost of Bellman backups.  
In this work, we demonstrate that the PWLC property allows a POMDP's value function to be alternatively represented as a finite set of neural networks.  
This insight enables a novel POMDP planning algorithm called \emph{Neural Value Iteration}, which combines the generalization capability of neural networks with the classical value iteration framework.  
Our approach achieves near-optimal solutions even in extremely large POMDPs that are intractable for existing offline solvers.  
\end{abstract}

\section{Introduction}

The partially observable Markov decision process (POMDP) framework models decision-making problems where the true state is hidden and transitions are stochastic.
In a POMDP, the agent maintains a \emph{belief} (a probability distribution over states) and seeks an optimal policy mapping beliefs to actions to maximize expected return.
However, solving POMDPs optimally is computationally intractable: PSPACE-complete in the finite-horizon case \citep{papadimitriou1987complexity} and undecidable for infinite horizons \citep{madani1999undecidability}.


It has been proved that the POMDP's optimal value function \( V^*_t \) at each time step \( t \) is piecewise linear and convex (PWLC) \citep{sondik1971optimal}.  
As a result, \( V^*_t \) can be efficiently represented by a finite set of \( |S| \)-dimensional hyperplanes, known as $\alpha$-vectors, where \( S \) denotes the state space.  
By leveraging this PWLC structure, value iteration methods for POMDPs (e.g., dynamic programming updates) can be performed by iteratively maintaining and updating a finite set of $\alpha$-vectors \citep{Pineau-ijcai03, Smith-uai04, Smith_HSVI2, sarsop, bai2010monte}.
However, since each $\alpha$-vector spans the entire state space, these methods are difficult to scale due to memory and computation costs.
Even efficient solvers like SARSOP \citep{sarsop} struggle with domains containing millions of states, where optimizing large sets of $\alpha$-vectors becomes intractable.
To improve scalability in continuous-state POMDPs, MCVI \citep{bai2010monte} implicitly represents $\alpha$-vectors using nodes in a policy graph (i.e., a finite-state controller) and performs approximate Bellman backups via Monte Carlo (MC) simulations.
Despite its scalability advantages, MCVI suffers from performance degradation in large problems with long horizons due to the high cost of MC simulations, limiting its practicality unless techniques like macro-actions \citep{macro_action_mcvi} are used to reduce planning depth and computational overhead.
\begin{figure}[t]
    \centering
    \includegraphics[width=0.7\linewidth]{ 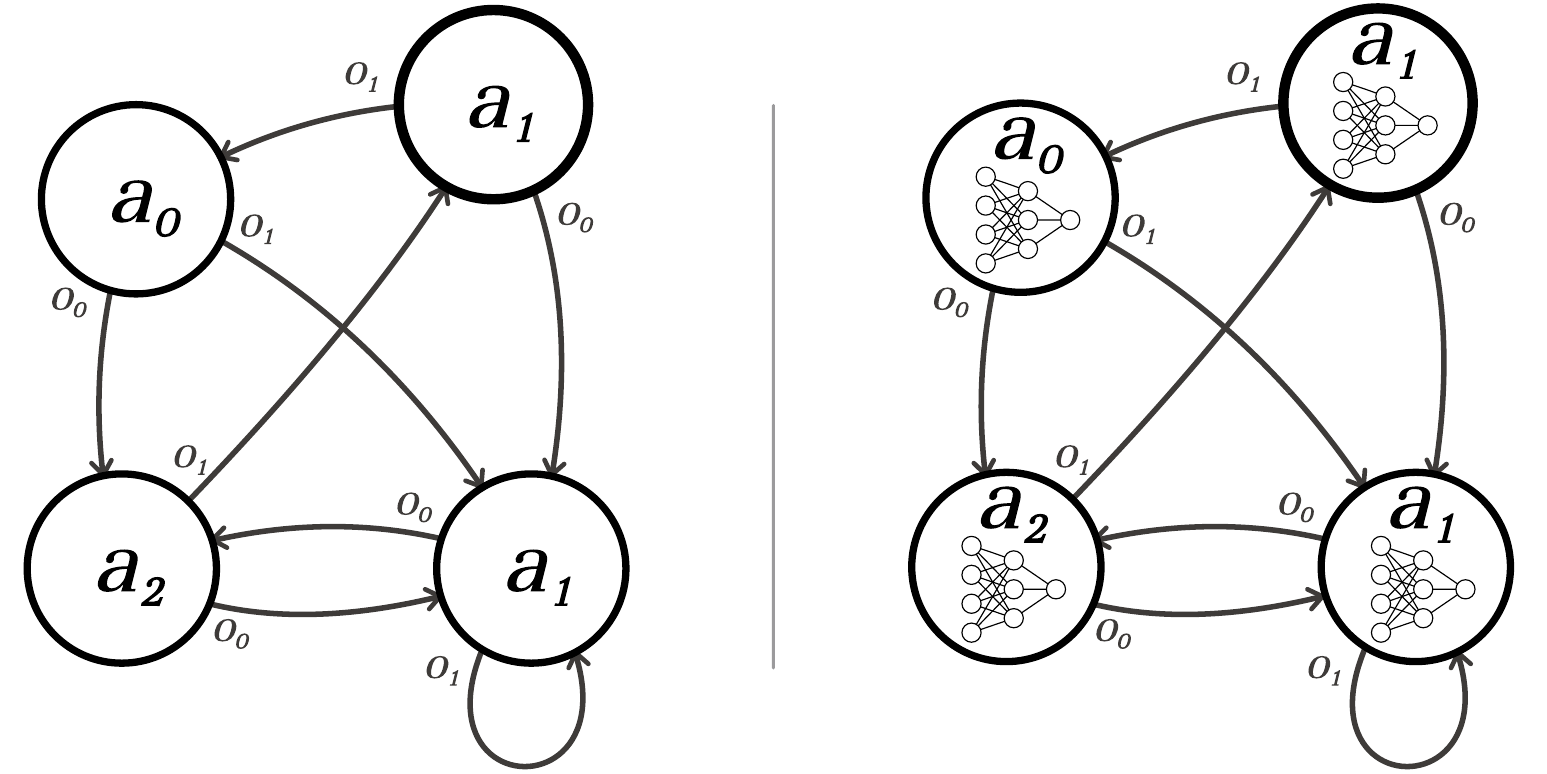}
\caption{A finite-state controller (FSC) vs. a finite-network controller (FNC).
Each FNC node stores a neural network that explicitly approximates an $\alpha$-vector.}
    \label{fig:fsc_vs_fnc}
\end{figure}

In this paper, based on the classic value iteration (VI) scheme, we aim to further extend its scalability to solve very large POMDPs.
To do so, we address a fundamental question: \emph{Is there a better way to represent a value function that is more suitable for such large-state POMDPs?}  
Motivated by the answer to this question, we propose a new perspective: one can leverage the PWLC property and efficiently approximate a POMDP's value function using a finite set of neural networks.  
This insight leads to a new policy representation, the \emph{finite-network controller} (FNC), where each node holds a neural network that explicitly models an $\alpha$-vector (\cref{fig:fsc_vs_fnc}), allowing compact policies even in complex domains.
We then introduce \emph{Neural Value Iteration} (NVI), a new offline solver that performs value iteration directly over these networks.
To our knowledge, NVI is the first offline POMDP solver to scale to problems as large as RockSample(20, 20), with over 400 million states—far beyond the capability of existing general offline POMDP solvers.


\section{Related Work}

POMDP planning techniques can be broadly categorized into offline and online methods.
Offline solvers aim to compute a complete solution prior to execution, typically in the form of a compact policy structure that can be readily executed at runtime.
In contrast, online planners interleave planning and execution by computing the best action for the current belief at each decision step, often using lookahead search or sampling techniques.
In this section, we mainly review the offline POMDP approaches that are most relevant to our contributions.
For comprehensive surveys and broader discussions, we refer the reader to existing literature \citep{lauri2022partially,kurniawati2022partially}.

Most offline POMDP solvers \citep{Smith-uai04,Smith_HSVI2,sarsop} rely on iteratively updating the value function via Bellman backups on a finite set of $\alpha$-vectors, which represent the POMDP value function.
All these methods and variants \citep{bai2010monte, macro_action_mcvi} can be viewed as following a \textit{bottom-up} approach: to solve a POMDP of horizon \( h \), they recursively construct the optimal value function backward in time, from \( t = h \) to \( t = 1 \).
Modern methods such as HSVI and SARSOP show that it is efficient to perform Bellman backups over a relatively small belief space by only sampling representative beliefs via reachability heuristics,
which largely speeds up value iteration convergence.
However, for very large POMDPs, even these value iteration methods are considered intractable.
An exception among recent offline methods is POMCGS \citep{you2025}, which does not follow a value iteration and instead uses a \textit{top-down} strategy inspired by online MCTS-based planners \citep{pomcp,despot,pomcpow,adaops,UCT}.
It treats the POMDP as a belief-MDP and performs tree search directly in belief space, merging similar nodes by comparing belief differences across depths to compact the policy tree into a graph and reduce redundant computation.
Its Monte-Carlo search framework also supports techniques such as action progressive widening, enabling efficient planning in continuous domains.

On the other hand, approximation techniques have been extensively studied to scale POMDPs to large or continuous domains.
A common approach is to approximate beliefs or value functions using parametric forms, particularly Gaussian mixtures \citep{thrun1999monte,spaan2005perseus,porta2006point,prentice2010belief}.
However, such representations often suffer from oversmoothing and struggle with discontinuities (e.g., walls or obstacles), requiring many Gaussian components to achieve pratical approximation, which prevents such methods from being applied on more complex domains.
Similar ideas have emerged in the reinforcement learning (RL) community.
In POMDP settings, RL-based methods \citep{drqn,ppo,karkus2017qmdp} often use recurrent neural networks (RNNs) to encode a hidden state as a belief surrogate.
The value function is then approximated by a neural network over this hidden representation.
Recent methods like BetaZero \citep{moss2023betazero} learn value functions over simplified belief statistics (e.g., mean and variance), using them as heuristics in MCTS planning over belief-MDPs.
However, none of these approaches explicitly leverage the piecewise-linear and convex (PWLC) structure of the POMDP value function, limiting their alignment with value iteration theory and associated guarantees.

In this work, we offer a new perspective: the PWLC structure of the optimal value function can be preserved by representing each $\alpha$-vector with a neural network.
This insight leads us to propose the \textit{finite-network controller} (FNC) and a corresponding offline algorithm, \textit{Neural Value Iteration} (NVI), that scales to POMDPs with large state spaces.


\section{Background}
\subsection{POMDPs}
A POMDP defines a single-agent decision-making problem in an environment with uncertain dynamics where the underlying states cannot be directly observed.
A POMDP is defined by a tuple $\langle \cS, \cA,\Omega,T,O,r,b_{0}\rangle$ where
  \begin{itemize}
  \item $\cS$ is the set of states;
  \item $\cA$ is the set of actions;
  \item $\Omega$ is the set of observations;
  \item $T(s',s,a) = \text{Pr}(s' | s, a)$ is the transition function, which encodes the dynamics of the environment (the probability of reaching the next state $s'$ given the current state $s$ and an executed action $a$);
  \item $O(o, s', a) = \text{Pr}(o | s', a)$ is the observation function; it indicates the probability of receiving an observation $o$ given the next state $s'$ reached while performing action $a$;
  \item $r(s,a)$ is the reward function, which gives the instant reward when performing action $a$ at state $s$;
  \item $b_{0}$ is an initial probability distribution over states.    
  \end{itemize}

In a POMDP, the agent maintains a belief $b$ at each time step which represents the probability distribution of possible current states.
When the agent executes an action $a$ and receives a new observation $o$, the agent updates its belief as $b' = \tau(b,a,o)$, where 
\begin{align}
    b'(s') = \frac{O(o,s',a)\sum_{s \in S}T(s',s,a)b(s)}{\sum_{s'' \in S} \left[ O(o,s'',a)\sum_{s \in S}T(s',s,a)b(s)\right] }
\end{align}
To simplify the expression, we use $\tau(b,a,o)$ to refer to the belief update in following sections.
The goal of a POMDP is to compute a policy $\pi$ that maps beliefs to agent actions that maximize the expected discounted accumulated reward.

\subsection{Alpha Vectors and Finite-State Controller}

A POMDP's value function \( V \) can be represented by a finite set of \( \alpha \)-vectors, denoted as $\Gamma$, i.e., \( \Gamma = \{\alpha_1, \dots, \alpha_n\} \), where each \( \alpha \in \Gamma \) is associated with a specific action.
Given a belief \( b \), the value function can be computed as:
\begin{align}
    V(b) = \max_{\alpha \in \Gamma} \sum_{s \in S} b(s)\, \alpha(s).
\end{align}
The optimal action for belief \( b \) is determined by the action associated with the maximizing \( \alpha \)-vector.\\
It is well known that a value function represented with a finite set of $\alpha$-vectors can be equivalently encoded as a finite-state controller (FSC), or policy graph, where each controller node corresponds to an $\alpha$-vector and is labeled with an action and a transition rule based on observations \citep{hansen1997improved}.
This transformation enables compact policy representations and efficient execution via table lookups, especially when the size of the FSC is small relative to the belief space.

\begin{definitionbox}{Finite-State Controller}{}
A (deterministic) \textit{Finite-State Controller} for a POMDP is a tuple \( \fsc = \langle N, n_0, \psi, \eta \rangle \), where:
\begin{itemize}
        \item \( N \) is a finite set of controller nodes (also called machine states),
    \item \( n_0 \in N \) is the initial controller node,
    \item \( \psi: n \rightarrow A \) is the action selection function, mapping each node to an action in the action space \( A \),
    \item \( \eta: N \times \Omega \rightarrow N \) is the node transition function, mapping a controller node and an observation \( o \in \Omega \) to the next node.
\end{itemize}
At each time step, the FSC selects an action based on the current node using \( \psi \), and upon receiving an observation, it updates the node using \( \eta \). The controller thus defines a reactive policy over partial observations.
\end{definitionbox}

\subsection{Bellman Backups in POMDPs}
In this section, we describe the core Bellman backup operation that underpins our contribution. For a broader overview of point-based methods, we refer the reader to \cite{shani2013survey}.
A POMDP's optimal value function can be computed through value iterations, and each iteration is commonly referred to as a \textit{Bellman Backup}.
This backup process updates a new value function $V'$ from the current value function $V$ as:
\begin{align}
    & V'(b) = \max_{a \in A} \left[ R(b,a) + \gamma \sum_{o \in \Omega}\text{Pr}(o|b,a)V(b^{a,o}) \right] \label{eq:raw_vi}\\
    & =  \max_{a \in A} \left[ R(b,a) + \gamma \sum_{o \in \Omega}\text{Pr}(o|b,a)\max_{\alpha \in \Gamma}\sum_{s'\in S}b^{a,o}(s')\alpha(s')\right] \label{eq:backup}.
\end{align}
%
The complexity of a raw backup computation in \Cref{eq:raw_vi} is $O(|V|\times|A|\times|\Omega| \times |S|^{2} + |A| \times |S|\times|V|^{\Omega})$, which means this kind of naive value iteration can only be done in very small problems with very few horizons.
One of the key breakthroughs in offline POMDP planning is the development of \emph{point-based value iteration} (PBVI) methods, which demonstrate that full Bellman backups over the entire belief space, as required by the raw value iteration in \Cref{eq:raw_vi}, are unnecessary.
Instead, PBVI methods maintain a finite set of representative beliefs $\mathcal{B}$ and reuse cached $\alpha$-vectors for efficient computation as shown in \Cshref{alg:bellman_backup}.
This compact backup operation generates a new $\alpha$-vector $\alpha'$ for a given belief $b$, and add it to the $\alpha$-vector set $\Gamma \gets \Gamma \cup \{\alpha'\}$, which represents the updated value function $V'$.

\begin{algorithm}[b!]
\caption{Bellman Backup}
\label{alg:bellman_backup}
\SetKwFunction{Backup}{Backup}
\SetKwProg{Fn}{Function}{:}{}
\Fn{\Backup{$\Gamma, b$}}{
    \ForEach{$a \in A$ \textbf{and} $o \in \Omega$}{
        $\alpha_{a,o} \gets \arg\max_{\alpha \in \Gamma} \left( \alpha \cdot \tau(b, a, o) \right)$
    }
    \ForEach{$a \in A$ \textbf{and} $s \in S$}{
        $\alpha_a(s) \gets R(s,a) + \gamma \sum_{o \in \Omega} \sum_{s' \in S} T(s,a,s')\, O(s',a,o)\, \alpha_{a,o}(s')$
    }
    $\alpha' \gets \arg\max_{a \in A} \left( \alpha_a \cdot b \right)$\;
    \Return $\Gamma \cup \{\alpha'\}$
}
\end{algorithm}

Nonetheless, when the state space is large or continuous, even this backup process becomes intractable due to the need to sum over all possible states. To address this challenge, Monte Carlo Value Iteration (MCVI) \cite{bai2010monte} was proposed. MCVI avoids explicit loops over the entire state space by sampling states and implicitly representing $\alpha$-vectors using policy graph nodes. It approximates the Bellman backup using Monte Carlo simulations, a procedure referred to as \emph{MC-Backup}.
However, the lack of explicit $\alpha$-vector representations in MC-Backup prevents caching and reusing computation across belief points. As a result, MCVI often requires a large amount of simulation to achieve competitive performance, which limits its solution quality in large-scale problems under constrained computational budgets.  

\section{Neural Value Iteration}

This section presents our main contribution.
At a high level, our algorithm maintains and trains a finite set of neural networks during planning to approximate the POMDP value function over a set of representative beliefs. The training data for each network comes from sampling outcomes.

\subsection{Represent $V$ with a Finite Set of Neural Networks}
\label{sec:FNC}

As discussed in the related work section, one natural approach to handling large-scale problems is to directly approximate the value function as a parametric mapping from the belief space $\mathbb{B}$ to a scalar value in $\mathbb{R}$. 
However, this approximation first requires the belief inputs to be some sort of simplified structure such as means and variances \citep{moss2023betazero}, or RNN's hidden states \citep{drqn,ppo}, which may lose some information compared with the original belief.
%
%
Additionally, updating the approximation for new beliefs may change its output for previously seen ones, introducing instability and reducing accuracy, especially when generalizing to unseen beliefs.

In this paper, we propose to revisit the key PWLC structure exploited by previous value iteration methods, which allows a POMDP value function $V$ to be represented as the maximum over a finite set of $\alpha$-vectors such that $V(b) = \max_{\alpha \in \Gamma} b \cdot \alpha$.
We then ask one fundamental question: What is an $\alpha$-vector mathematically?
A simple but important answer is that each $\alpha$-vector is a mapping from $\mathbb{S}$ to $\mathbb{R}$.
This means that we can learn it with a function approximator (\eg, neural network) $\alpha_{\text{NN}}$, and then we can represent a POMDP value function as the maximum output of a set of neural networks such that $V(b) = \max_{\alpha_{\text{NN}}} b \cdot \alpha_{\text{NN}}$.
This representation allows efficient approximation of the value function over large or continuous state spaces, and also provides direct access to state-wise values through $\alpha_{\text{NN}}(s)$ with $ \forall s \in S$.
Moreover, we can further define a new policy type called a \textit{finite network controller} (FNC) as follows:

\begin{definitionbox}{Finite Network Controller}{}
    A \textbf{\textit{Finite Network Controller}} (FNC) is a variation of a finite state controller with $G = \langle N, n_0,\psi, \eta \rangle$, which has the following features:
    \begin{enumerate}
        
        \item In addition to the labeled best action, each node $n \in N$ in an FNC also stores a neural network $\alpha_{\text{NN}}$ to explicitly represent an $\alpha$-vector;
        \item The value of a belief $b$ in an FNC denoted as $G$ can be directly computed as $V(b) \gets \max_{n \in G}b\cdot n.{\alpha_{\text{NN}}}$;
        \item An FNC can be accurately transformed to an FSC by removing the neural network stored in each node.
    \end{enumerate}
\end{definitionbox}

This new type of policy allows us to facilitate the value iteration process (as shown in the next section, \Cref{alg:bellman_backup}) with sampling-based simulators, and also enables maintaining a compact structure during execution by transforming the FNC into a classic FSC without any performance loss by simply removing the neural network stored in each node.

\subsection{Value Iteration with Neural Bellman Backups}
\label{sec:NBB}

In this section, we present our main contribution: \textit{Neural Value Iteration} (NVI), which performs value iteration (VI) with a finite network controller for large-scale POMDPs.
Such domains are often accessible only through black-box simulators, requiring Monte Carlo sampling for computations.
Therefore, we adopt MCVI as the base VI framework and show how our approach integrates into it.
The overall procedure for performing value iteration over neural networks is outlined in Algorithm~\ref{alg:main},
which follows a classical VI scheme: iteratively collecting beliefs and performing backups at those belief points until convergence.

We use a standard belief collection method. Briefly, $\textit{CollectBelief}$ performs a forward tree search according to a reachable belief heuristic, where actions are selected according to the highest Q-values and observations are selected to maximize the uncertainty between child beliefs' upper and lower bounds. The beliefs in this search trajectory are collected. In our case, the belief update is a particle filtering process, where each belief $b$ is represented by a collection of $nb_{\text{particle}}$ states. 
A more detailed belief collection algorithm is provided in the appendix, as it is not our main contribution.

The key part of our contribution is presented in Algorithm~\ref{alg:neural_bellman_backup}. This \textit{Neural Bellman backup} method defines how to compute a new value function $V'$ for a given belief $b$ in a continuous-state problem:
\begin{equation}
\begin{aligned}
\label{eq:backup_continuous}
    V'(b) &= \max_{a \in A} \bigg\{  \textcolor{blue}{ \int_{s \in S} R(s,a)\, b(s)\, ds }
     + \textcolor{brown}{  \gamma \sum_{o \in \Omega} \text{Pr}(o|b,a)\max_{\alpha \in \Gamma} \int_{s'\in S} b^{a,o}(s')\, \alpha(s')\, ds' } \bigg\}
\end{aligned}
\end{equation}
Algorithm~\ref{alg:neural_bellman_backup} performs a Bellman backup on a given belief $b$ and updates a finite-network-controller (FNC). It starts with an initialization process (Lines~\ref{algline:init_ra}-\ref{algline:init_V}), and then computes the expected immediate reward (marked in blue in Equation~\ref{eq:backup_continuous}) with simulations ( Lines~\ref{algline:loop_states}-\ref{algline:compute_ra}).
A key difference compared with MCVI is how to compute the discounted future return (marked in brown in Equation~\ref{eq:backup_continuous}), particularly regarding $\alpha(s')$. Because it's challenging to explicitly represent $\alpha$-vectors for large or continuous domains, MCVI implicitly represents the value of $\alpha(s')$ by performing Monte Carlo simulations through an FSC policy starting from state $s'$. This computation becomes very expensive when the problem and FSC size are large. Moreover, to get an accurate approximation of future rewards with stochastic dynamics, MCVI typically requires a large number of simulations.\\
In our case, we explicitly represent each $\alpha$-vector with a neural network $\alpha_{\text{NN}}$ stored at each node, and use it to predict the value of $\alpha(s')$ in Line~\ref{algline:predict_value_s}, which significantly reduces computations. Lines~\ref{algline:loop_obs}-\ref{algline:new_node} then compute the discounted expected future return, and we create a new temporary node with the best action found. Different from MCVI, we add a pruning process before adding the new node to $G$. This process (function $\textit{NodeIsUnique}$ in Algorithm~\ref{alg:neural_bellman_backup}) checks if there exists a node in $G$ with the same best action and node transition edges.

When adding a new node $n'$, a key process is to assign the new node a neural network $\alpha_{\text{NN}}$ that explicitly represents an $\alpha$-vector. We propose to use a separate function $\textit{GenData}$ to generate training data for $\alpha_{\text{NN}}$. To collect representative states so that $\alpha_{\text{NN}}$ can generalize to unseen states, this function first randomly samples $nb_{\text{sample}}$ different states from the environment with all possible time depths (Lines~\ref{algline:loop_s_gendata}-\ref{algline:get_s_gendata}). These $nb_{\text{sample}}$ different states are used as the training dataset $X$.\\
Then for each collected state sample, if the current FNC $G$ is empty, we compute $V_s$ by simulating $nb_{\text{sim}}$ rollouts of always performing the same action $n'.a^*$ (Line~\ref{algline:Vs_emptyG}), as there are no transitions between nodes. If $G$ is not empty, we calculate $V_s$ with expected immediate rewards plus the discounted future return $V_{n^{(i)}, s'^{(i)}}$ as shown in Line~\ref{algline:Vs_neural}, where $s'^{(i)}, o^{(i)}, r^{(i)} \gets \text{Env}_{\text{sim}}(s, n'.a^*)$ and $n^{(i)} = \eta(G, n', o^{(i)})$.
This computation is straightforward because $V_{n^{(i)}, s'^{(i)}}$ simply calls the neural network stored in node $n^{(i)}$ to predict a value such that $V_{n^{(i)}, s'^{(i)}} = n^{(i)}.\alpha_{\text{NN}}(s'^{(i)})$. Finally, all collected $V_s$ values serve as training labels $Y$. Together, $X$ and $Y$ are used to train $\alpha_{\text{NN}}$ and store it in the new node $n'$, and we update the FNC $G$ by adding this new node $n'$.

In practice, data collection and training for each $\alpha_{\text{NN}}$ are quite efficient, as only a relatively small dataset is needed per node.
A simple neural network architecture such as a multilayer perceptron (MLP) with one or two hidden layers is typically sufficient for most POMDP problems.
From an engineering perspective, once $nb_{\text{sample}}$ states are collected to approximate the state space, they can be stored and reused in subsequent $\text{GenData}$ operations. Furthermore, the value prediction step in \cref{algline:predict_value_s} can be optimized by batching the evaluation of multiple states, significantly improving computational efficiency.

Algorithm~\ref{alg:neural_bellman_backup} can be seen as a sampled approximation of Algorithm~\ref{alg:bellman_backup}, where exact computations over the full state space are replaced by Monte Carlo simulations. 
More importantly, it uses neural networks $\alpha_{\text{NN}}$ to handle large or continuous states instead of the original $\alpha$-vectors, and the value function is bootstrap-updated with a finite set of neural networks.

\subsection{Benefits of $\alpha$-vector Recasting}

We analyze the detailed benefits of recasting $\alpha$-vectors with neural networks.

A significant advantage is the exponentially reduced number of simulator calls compared to MCVI. 
Specifically, in each MCVI backup, since the algorithm does not maintain an explicit $\alpha$-vector representation, the value $\alpha(s')$ (or $V_{n,s'}$ in \Cref{algline:predict_value_s}) must be estimated via Monte-Carlo simulations using the current policy starting from node $n$ and state $s'$. To obtain a reliable estimate, a large number of rollouts ($nb_{\text{rollout}}$) and considerable simulation depth ($d$) are typically required. Consequently, each MCVI backup requires looping over actions ($a \in A$), belief particles ($s \in b$), and nodes ($n \in G$), demanding approximately 
\[
|A| \times nb_{\text{particle}} \times nb_{\text{node}} \times nb_{\text{rollout}} \times d
\] 
simulator calls to access all $\alpha(s')$. Even with a fast simulator, this cost remains high and grows with the size of the node set. More critically, for larger or highly stochastic problems, the required number of belief particles $nb_{\text{particle}}$, simulation depth $d$, and rollouts $nb_{\text{rollout}}$ must increase further to ensure accurate Monte‑Carlo evaluation; otherwise, performance degradation may occur as observed in \cite{macro_action_mcvi}.

In contrast, like other offline solvers that maintain explicit $\alpha$-vector representations, NVI can directly access the value $\alpha(s')$ and, more importantly, predict $\alpha(s')$ in batches with negligible time. 
This capability reduces the required simulator steps by \emph{orders of magnitude} compared to MCVI, yielding substantial computational savings even when the simulator is fast. One might argue that the burden is merely shifted to the $\text{GenData}$ process and neural network training. However, as explained in the previous section, only the first NVI backup (when $G$ is empty) requires full rollouts, and those rollouts loop only over $nb_{\text{sample}}$ states—not over actions and each belief particle as in MCVI. Moreover, each NVI backup trains only a small neural network (e.g., an MLP) using $nb_{\text{sample}} \ll |S|$ samples, making the training load very light and easily accelerated on modern GPUs.

A further benefit emerges when comparing NVI to classic offline solvers such as SARSOP or HSVI, which store $\alpha$-vectors explicitly. For large problems like RockSample($20, 20$) with a state space $|S| > 400$M, even minimal storage per vector exceeds 14GB, rendering such approaches memory‑infeasible. By contrast, a 3‑layer MLP (with 1024 neurons per layer) representing $\alpha_{\text{NN}}$ requires $<20$MB.

These observations precisely identify the scalability barriers that have limited prior offline solvers. Sampling‑based methods like MCVI can handle small‑ to medium‑scale problems, but their reliance on costly Monte‑Carlo simulations restricts their practicality in large, long‑horizon domains \cite{macro_action_mcvi}; obtaining a near‑optimal policy in such settings may demand unacceptable computation time. Solvers that store $\alpha$-vectors explicitly, such as SARSOP and HSVI, can deliver $\epsilon$-optimal solutions for small and certain medium‑size problems. However, without exploiting state‑gradient information, their memory requirements for storing $\alpha$-vectors explicitly become prohibitive for large state spaces.

NVI addresses these limitations by leveraging state gradient information through the representation of $\alpha$-vectors as $\alpha_{\text{NN}}$, making $\alpha$-vectors tractable in huge state spaces with acceptable fitting error. We argue that this learning for planning approach, where neural representations preserve the state-gradient structure of $\alpha$-vectors, is a crucial step toward scalable offline POMDP planning.

\begin{algorithm}[t!]
\caption{Neural Value Iteration}
\label{alg:main}
\SetKwFunction{Main}{Main}
\SetKwFunction{CollectBeliefs}{CollectBeliefs}
\SetKwProg{Fn}{Function}{:}{}

\Fn{\Main{$b_0, nb_{\text{particle}}, nb_{\text{sample}}, nb_{\text{sim}}$}}{

$G \gets \emptyset$ \\
\While{$V(b_0)$ Not Converged}
{
    $\mathcal{B} \gets \text{CollectBeliefs}(b_0, nb_{\text{particle}})$ \\
    \ForEach{ $b \in \mathcal{B}$}{
        $\text{NeuralBackUp}(G, b, nb_{\text{sample}}, nb_{\text{sim}})$
    }
}

\Return $G$
}
\end{algorithm}

\begin{algorithm}[t!]
\caption{Neural Bellman Backup}
\label{alg:neural_bellman_backup}
\SetKwFunction{NeuralBackUp}{NeuralBackUp}
\SetKwFunction{GenData}{GenData}
\SetKwProg{Fn}{Function}{:}{}
\Fn{\NeuralBackUp{$G, b, nb_{\text{sample}}, nb_{\text{sim}}$}}{

$\text{For each } a \in A, R_a \gets 0$ \label{algline:init_ra} \\
$\text{For each } a \in A, o \in \Omega, \text{and } n \in G, V_{a,o,n} \gets 0$. \label{algline:init_V} \\
\ForEach{$a \in A$}
{
\ForEach{$s \in b$ \label{algline:loop_states}}
{
    $s',o,r, \gets \text{Env}_{\text{sim}}(s,a)$ \\
    $R_a \gets R_a + r$ \label{algline:compute_ra} \\
    \ForEach{$n \in G$}
    {
        $V_{n,s'} \gets \text{Predict}(n.{\alpha'_{\text{NN}}}, s')$  \label{algline:predict_value_s} \\
        $V_{a,o,n} \gets V_{a,o,n} + V_{n, s'}$ \\
    }
  }
    \ForEach{$o \in \Omega$ \label{algline:loop_obs}}
    {
        $V_{a,o} \gets \max_{n \in G}V_{a,o,n}$ \\
        $n^*_{a,o} \gets \argmax_{n \in G}V_{a,o,n}$
    }
    $V_a \gets \frac{(R_a + \gamma \sum_{o \in \Omega}V_{a,o})}{|b|}$ \\
}
  $V^{*} \gets \max_{a \in A}V_a, \quad a^{*} \gets \argmax_{a \in A}V_a$ \\

 Create a new temporary node $n'$ with action $a^*$ and set edges $(n', n^*_{a^*,o})$ for each $o \in \Omega$\label{algline:new_node} \\  
\If{\text{NodeIsUnique}$(n', G)$} 
 {  
    $\alpha'_{\text{NN}}\gets \emptyset$ \\
    $X,Y \gets \text{GenData}(n', G, nb_{\text{sample}}, nb_{\text{sim}})$ \\
    $\text{TrainNN}(\alpha'_{\text{NN}},X,Y)$ \\
    Label new node $n'$ with network $\alpha'_{\text{NN}}$ \\
    $G \gets G \cup \{n'\} $ \\

 }   
    
    \Return $G$
}

\Fn{\GenData{$n', G, nb_{\text{sample}}, nb_{\text{sim}}$}}{

$X, Y \gets \emptyset$\\
\For{$i =1$ to $nb_{\text{sample}}$ \label{algline:loop_s_gendata}}
{
    $s \sim \text{Env}_{sim}$ \label{algline:get_s_gendata} \\
    $X \gets X \cup \{s\}$ \\
    $V_s \gets 0$ \\
    \eIf{$G$ is empty}
    {
    $V_s \gets \frac{\sum_{i=1}^{\text{nb}_{\text{sim}}} \left( \sum_{t=0}^{\infty} \gamma^t R(s_t^{(i)}, n'.a^*) \right)}{\text{nb}_{\text{sim}}}$ \label{algline:Vs_emptyG}\\
    }{
        $V_s \gets \frac{\sum_{i=1}^{\text{nb}_{\text{sim}}} \left( r^{(i)} + \gamma \cdot V_{n^{(i)}, s'^{(i)}} \right)}{\text{nb}_{\text{sim}}}$ \label{algline:Vs_neural} \\
    }
    $Y \gets Y \cup \{ V_s \}$ \\
}
\Return $X, Y$
}
\end{algorithm}

\section{Experiment}
We implement our main contribution, NVI, in Julia using the POMDP.jl framework \citep{pomdpjl}.
Planning experiments are conducted on a machine with an Intel i7 3.6GHz CPU, 64GB RAM, and an RTX 4060 GPU, while RL experiments are performed on an NVIDIA A100-SXM4-40GB GPU.

In NVI, each FNC node stores an MLP network as $\alpha_{\text{NN}}$ to explicitly represent an $\alpha$-vector. The network takes a vectorized state $s$ as input and outputs the estimated value $V_{n,s}$.
Additional experimental details are provided in the appendix.

\subsection{Benchmark Domains}
To evaluate our contribution's performance, we conduct experiments on the following benchmark domains from the POMDP.jl framework. These domains cover the most challenging problems for offline methods, including continuous POMDPs and discrete POMDPs with very large state spaces.

\paragraph{Light Dark}
The Light Dark problem \citep{pomcpow} involves an agent moving in a 1D world. The agent does not know its initial location and only receives accurate observations in a specific light area. The agent's goal is to localize itself and navigate to a target region.

\paragraph{Lidar Roomba}
The Lidar Roomba domain \citep{roomba} features a cleaning robot that must localize itself in a 2D environment and reach a goal position. The robot has no knowledge of its initial location and is equipped with a noisy lidar sensor that measures forward distance.

\paragraph{RockSample} 
The RockSample (RS) problem \citep{Smith-uai04} is defined by a tuple $\langle n,k \rangle$, where a robot moves in an $n \times n$ grid world containing $k$ rocks. Initially, the robot knows its own location but not the rocks' status (good or bad). The robot can measure rock quality with a noisy sensor whose accuracy depends on the distance to the rock. The goal is to collect as many good rocks as possible before reaching the target location.

Currently, one of the largest POMDP domains tested in recent literature is RS$(20,20)$, which contains over 419 million states. To our knowledge, no previous offline POMDP solver has successfully solved RS$(20,20)$, apart from our proposed NVI approach.


\begin{figure*}[t!]
\centering

\begin{subfigure}{0.32\textwidth}
  \includegraphics[width=\linewidth]{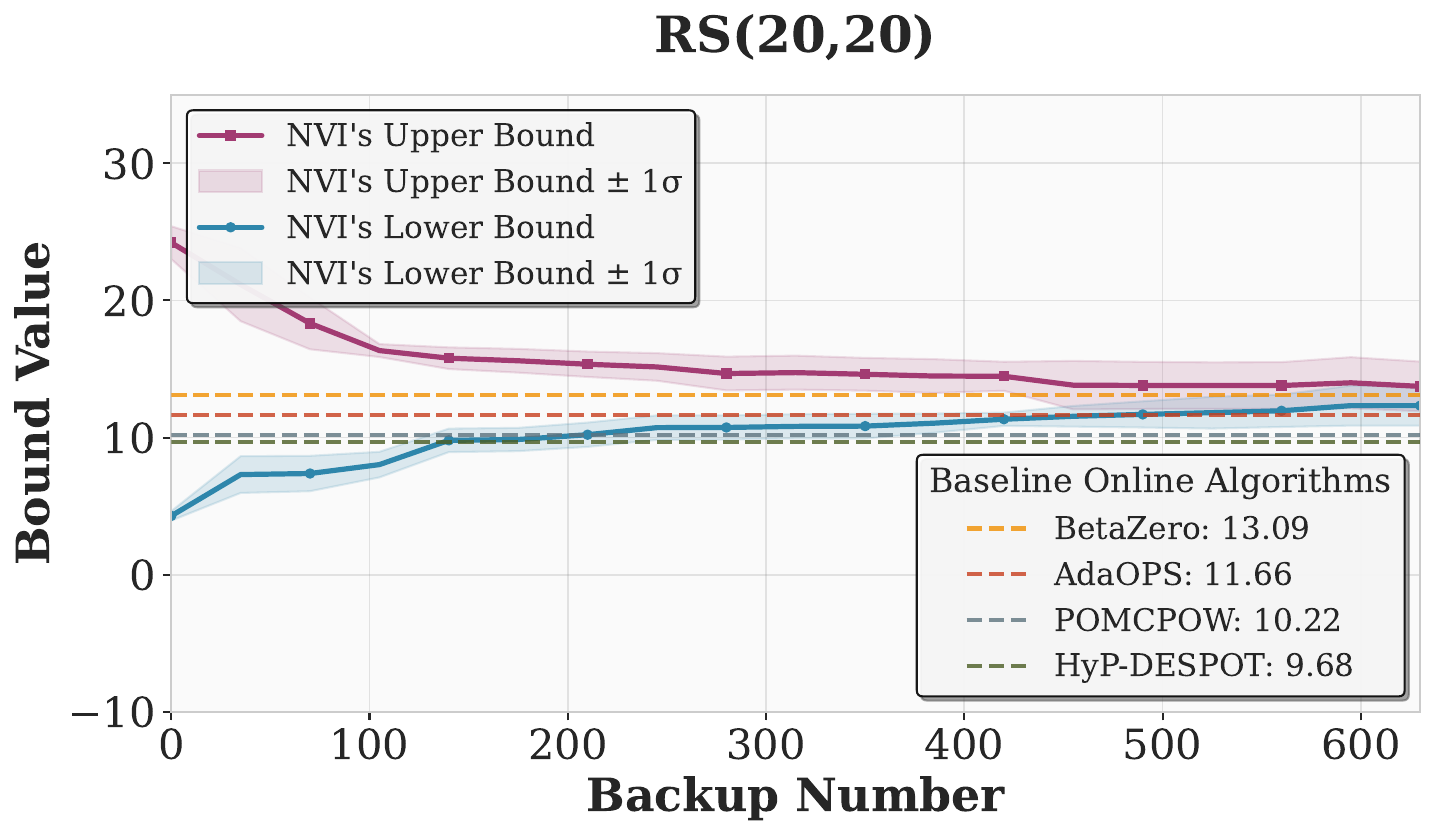}
  \caption{RS (20,20)}
  \label{fig:rs_bounds}
\end{subfigure}
\hfill
\begin{subfigure}{0.32\textwidth}
  \includegraphics[width=\linewidth]{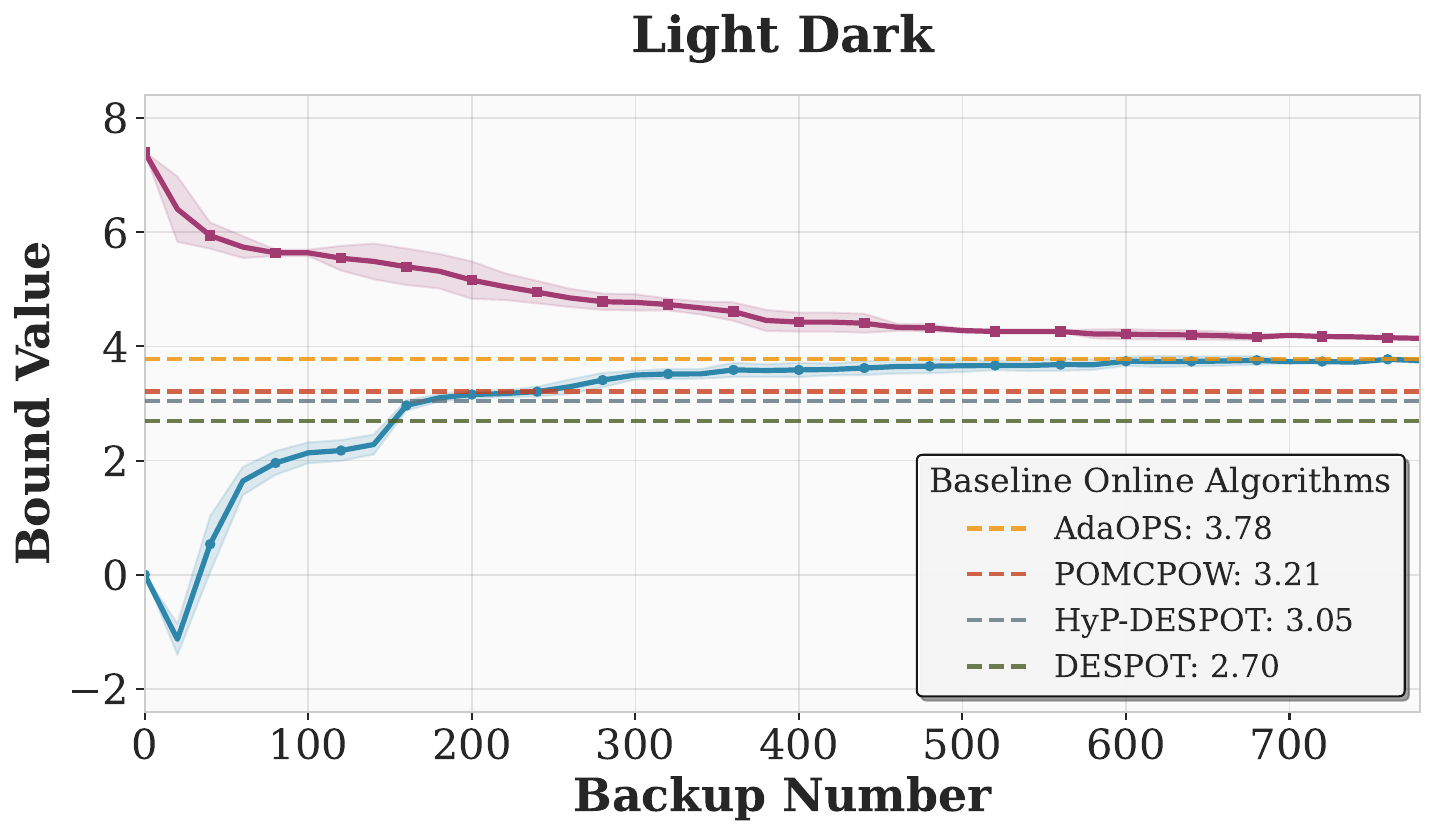}
  \caption{Light Dark}
  \label{fig:ld_bounds}
\end{subfigure}
\hfill
\begin{subfigure}{0.32\textwidth}
  \includegraphics[width=\linewidth]{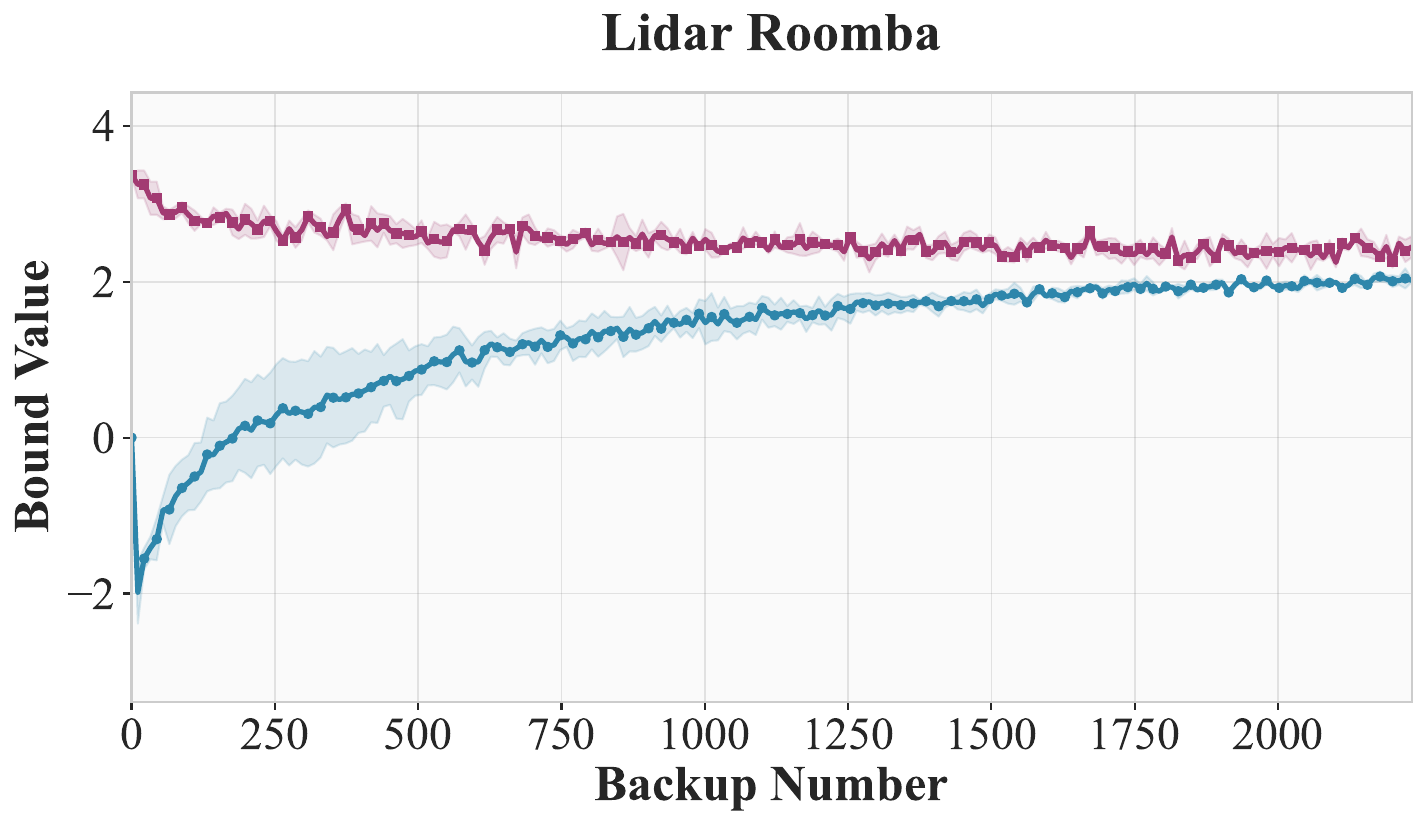}
  \caption{Lidar Roomba}
  \label{fig:lidar_bounds}
\end{subfigure}

\caption{NBB’s upper and lower bounds with backup numbers on RS (20,20), Light Dark, and Lidar Roomba. Baseline online planning methods \citep{despot, pomcpow, cai2021hyp, adaops} are reported from BetaZero \citep{moss2023betazero}.}
\label{fig:bounds_rs_ld}
\end{figure*}

\begin{figure*}[t!]
\centering

\begin{subfigure}{0.32\textwidth}
  \includegraphics[width=\linewidth]{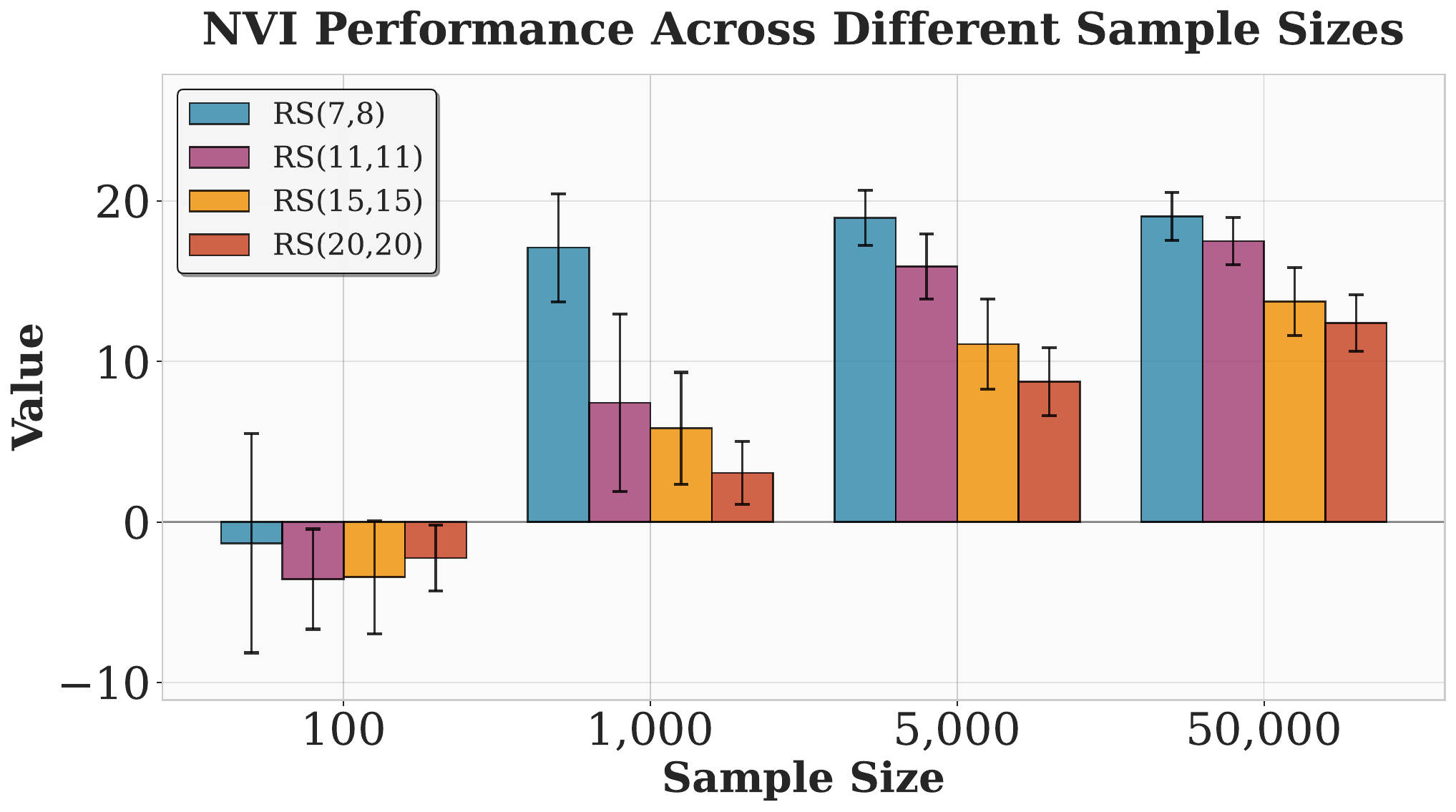}
  \caption{Impact of $nb_{\text{sample}}$ in RS domains}
  \label{fig:nbb_sample_bar}
\end{subfigure}
\hfill
\begin{subfigure}{0.32\textwidth}
  \includegraphics[width=\linewidth]{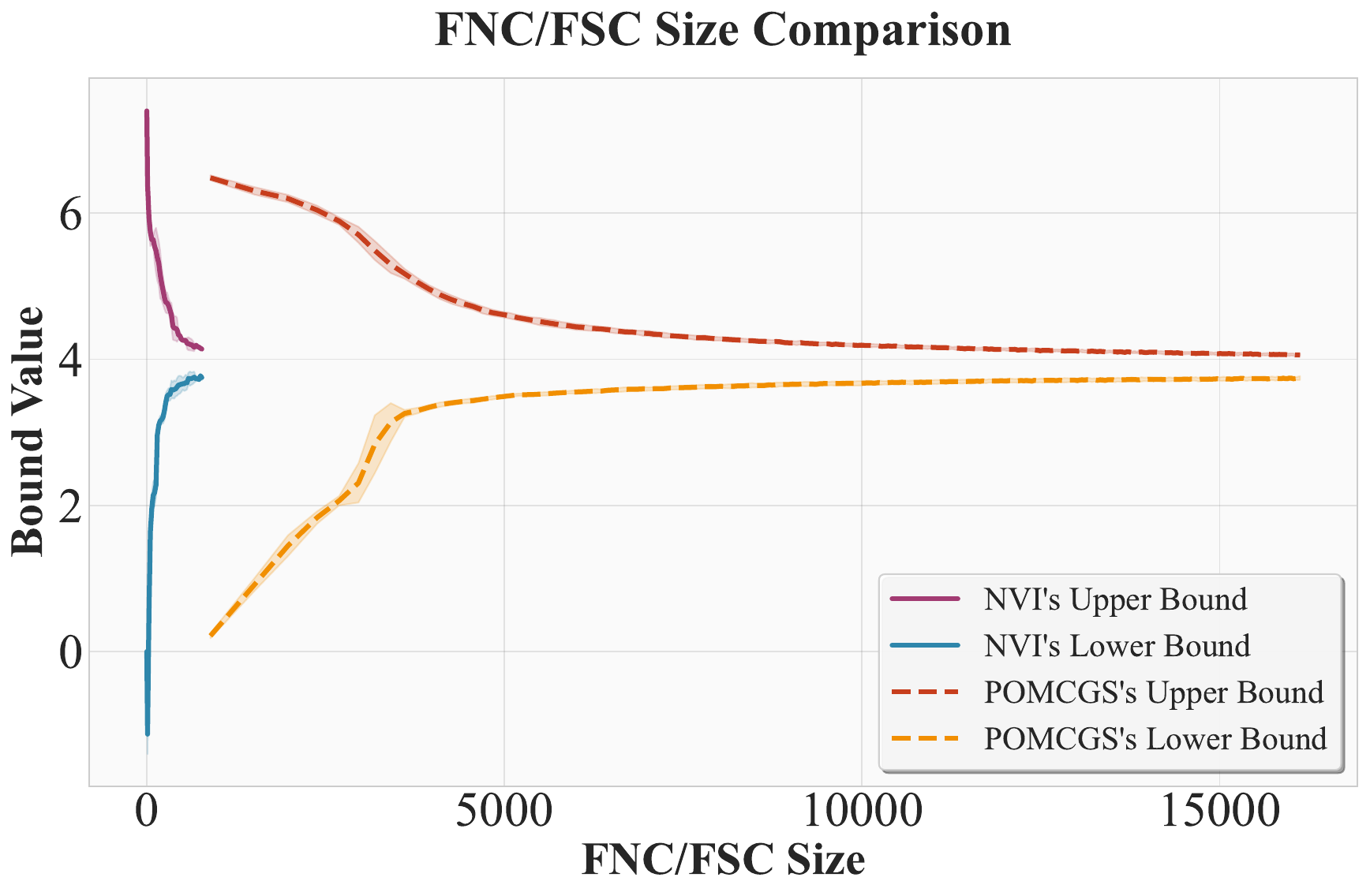}
  \caption{NVI vs POMCGS: FSC size}
  \label{fig:fsc_comparison}
\end{subfigure}
\hfill
\begin{subfigure}{0.32\textwidth}
  \includegraphics[width=\linewidth]{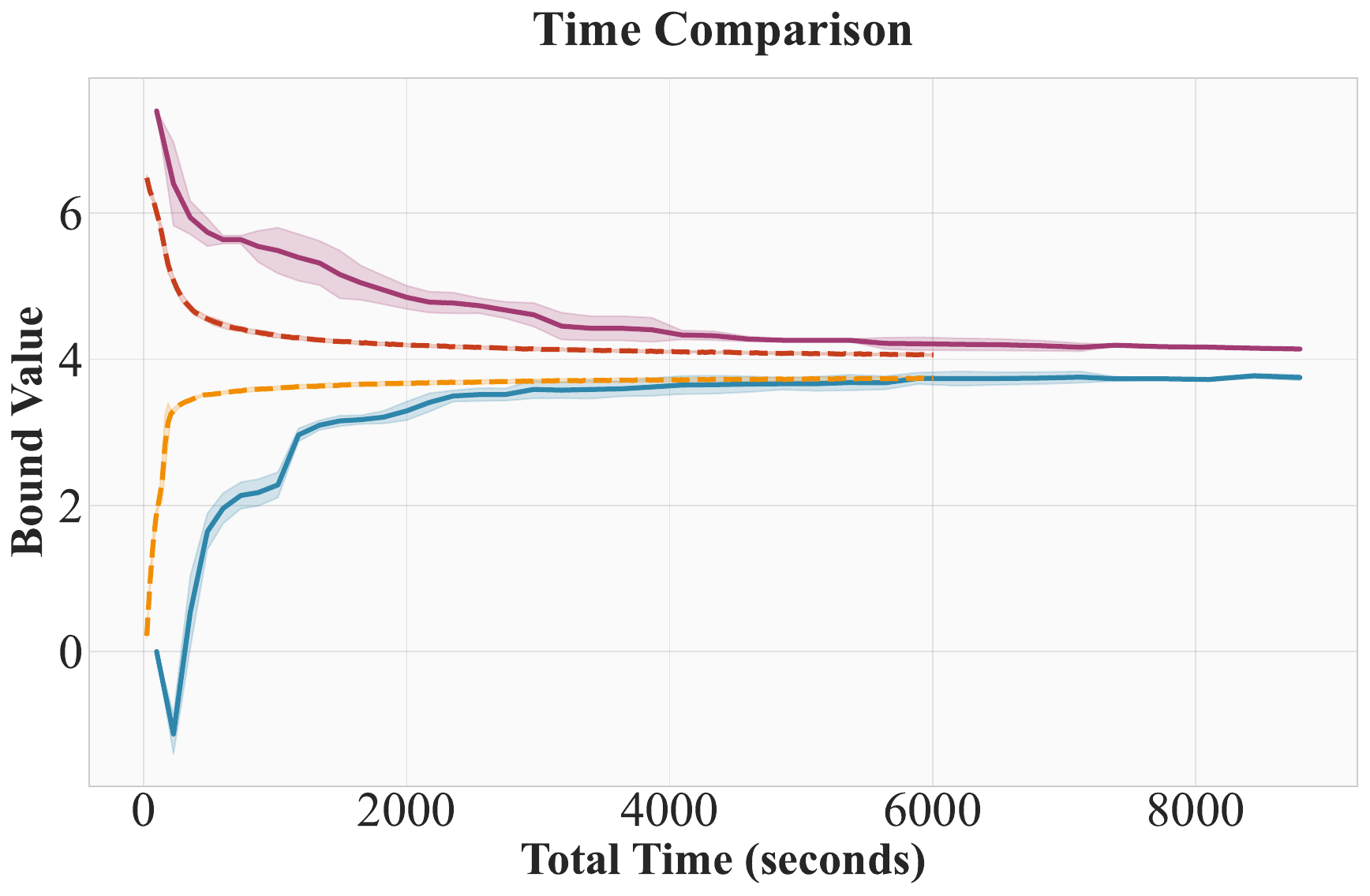}
  \caption{NVI vs POMCGS: planning time}
  \label{fig:time_comparison}
\end{subfigure}

\caption{NVI analysis: (a) Effect of $nb_{\text{sample}}$ on RS; (b, c) Comparison with POMCGS in FSC size and planning time on the Light Dark domain. NVI’s FNC is converted to a FSC.}
\label{fig:NBB_combined}
\end{figure*}

\subsection{Offline POMDP Methods}

\begin{table*}[t!]
\caption{\textbf{\textit{Performance comparison of different offline POMDP algorithms.}} -- The symbol ``$\dagger$" indicates out-of-memory errors during computation, and ``\text{---}" denotes algorithm inapplicability to the domain. All methods' performance is evaluated over 5 different runs, with sampling-based approaches' policies evaluated using $10^5$ simulations per run. In RockSample domains, each run uses a distinct configuration with randomized rock positions and statuses, while the robot always starts at position $(1,1)$ to maximize the distance to the exit area.
  }
  \centering
  \resizebox{1.0\linewidth}{!}{
    \begin{tabular}{lrrrrrr}
    \toprule
           & {RS$(7, 8)$} &  {RS$(11, 11)$} &  {RS$(15, 15)$} & {RS$(20,20)$} & Light Dark  & Lidar Roomba \\
    \midrule
    $|\cS|$ & $12,544$ & $247,808$ & $7,372,800$ & $419,430,400$ & $\infty$ & $\infty$\\
    $|\cA|$ & $13$ & $16$ & $20$ & $25$ & $3$ & $\infty$\\
    $|\cZ|$ & $3$ & $3$ & $3$ & $3$ & $\infty$ & $\infty$ \\   
    \midrule
    Recurrent PPO & $7.73 \pm 0.01$ & $6.30 \pm 0.04$ &$6.16 \pm 2.88$ & $3.97 \pm 0.02$  & $3.26 \pm 0.01$ & $-2.07 \pm 0.02$  \\
    SARSOP* & $\mathbf{21.57 \pm 1.82}$ &  $\mathbf{19.09 \pm 2.40}$ & \text{---} & \text{---} & \text{---} & \text{---} \\
    POMCGS & $20.17\pm 1.68$ & $18.53 \pm 1.04$ & $13.45 \pm 1.66$ & $3.91 \pm 0.83$ $(\dagger)$& $\mathbf{3.74 \pm 0.06}$ & $1.08 \pm 0.19$ \\ 
    MCVI  & $17.34 \pm 2.52$ & $13.78 \pm 2.13$ & $10.78 \pm 1.92$ & $4.57 \pm 2.19$ & $2.94 \pm 0.15$ & $0.68 \pm 0.16$  \\
    \textbf{NVI}  & $18.95 \pm 1.49$ & $17.51 \pm 1.56$ & $\mathbf{13.60 \pm 2.27}$ & $\mathbf{12.31 \pm 1.65}$ & $3.73 \pm 0.10$ & $\mathbf{2.03 \pm 0.03}$   \\
    \bottomrule
    \end{tabular}
  }
  %
\label{table:performance}
\end{table*}

We compare NVI with several state-of-the-art offline POMDP methods including SARSOP, MCVI, and POMCGS, along with a reinforcement learning baseline (PPO with recurrent neural network). For all planning methods, we set time limits of:
\begin{itemize*}
    \item 1 hour for small problems (RS$(7,8)$);
    \item 6 hours for medium problems (RS$(11,11)$ and Light Dark);
    \item 24 hours for large problems (RS$(15,15)$ and RS$(20,20)$ and Lidar Roomba).
\end{itemize*}
No time limit is applied to PPO.
Among all tested algorithms, only SARSOP requires explicit models providing exact state transition and observation probabilities.
Note that model loading and algorithm initialization times (which can be substantial, e.g., many hours for SARSOP) are excluded from these limits. For continuous domains, all sampling-based offline solvers require discretization; the details are provided in Table~\ref{table:discretization}.

\paragraph{Parameter Analysis}

NVI's planning process involves three key parameters:
    $nb_{\text{sample}}$, which controls how many sampled states are used to approximate the full state space during the training of $\alpha_{\text{NN}}$;
    $nb_{\text{particle}}$, which determines how many particles are used to represent a belief; and
    $nb_{\text{sim}}$, which sets the number of Monte Carlo simulations performed per particle to estimate expected outcomes.
The parameters $nb_{\text{particle}}$ and $nb_{\text{sim}}$ are common to most Monte Carlo-based planners, both online and offline, as these methods typically rely on particle filtering and simulation for belief tracking and value estimation. Thus, we omit a detailed discussion of these two parameters and focus our analysis on $nb_{\text{sample}}$, which is only used in NVI among current offline approaches.

As shown in \Cref{fig:nbb_sample_bar}, increasing $nb_{\text{sample}}$ generally improves NVI’s performance. For smaller domains such as RS($7,8$), a relatively small $nb_{\text{sample}}$ suffices to train $\alpha_{\text{NN}}$ effectively and enable generalization to unseen states. If $nb_{\text{sample}}$ exceeds the size of the state space $|S|$, $\alpha_{\text{NN}}$ essentially memorizes the entire state space, behaving more like a tabular representation.
For larger domains, a higher $nb_{\text{sample}}$ is typically needed to achieve good generalization. However, even then, $nb_{\text{sample}} \ll |S|$, underscoring the efficiency of using neural networks to approximate $\alpha$-vectors.
%



\begin{table}[t!]
\caption{\textbf{\textit{Discretization approaches for sampling-based methods}} -- Type notation: C=Continuous, D=Discrete (state/action/observation). Numbers in parentheses indicate discrete sizes; ``---" denotes no discretization required.}
\centering
\vspace{2mm}
\begin{tabular}{@{}lll@{}}
\toprule
\textbf{Problem (Type)} & \textbf{MCVI \& NVI} & \textbf{POMCGS} \\
\midrule
\multicolumn{3}{@{}l}{\textbf{Light Dark (C,D,C)}} \\
\quad States & --- & State Grid $= [1.0]$ \\
\quad Actions & --- & --- \\
\quad Observations & KMeans (20) & KMeans (20) \\
\midrule
\multicolumn{3}{@{}l}{\textbf{Lidar Roomba (C,C,C)}} \\
\quad States & --- & State Grid (4002) \\
\quad Actions & Subset (9) & Progressive Widening \\
\quad Observations & KMeans (10) & KMeans (10) \\
\bottomrule
\end{tabular}
\label{table:discretization}
\end{table}

\paragraph{Performance Comparison}Performance results are shown in Table~\ref{table:performance}. In small to moderate-sized problems, SARSOP achieves the highest returns due to its exact model computations. However, even when SARSOP can solve RS$(11,11)$ with high scores, it remains impractical due to excessive memory requirements and long loading times. 
For example, SARSOP requires about 14GB RAM and a very long time (more than 24 hours) to load and initialize an RS$(11,11)$ domain.
These limitations, combined with the complexity of exact Bellman backups over the entire state space $|S|$, prevent SARSOP from scaling to larger problems.
On the other hand, sampling-based approaches demonstrate better scalability for large and continuous domains.
Notably, among these methods, only NVI successfully solves the extremely large RS(20,20) domain with over 419 million states. %
Although MCVI should theoretically achieve performance similar to that of SARSOP or NVI given sufficient Monte Carlo simulations and particle sizes, its lack of explicit $\alpha$ vector representation dramatically increases the computational requirements for accurate estimation in large problems.
Another interesting observation is that despite using simple discretization schemes for continuous domains (Light Dark and Lidar Roomba), sampling-based methods such as POMCGS and NVI perform relatively well on these problems. This suggests that continuous domains may not inherently pose significant challenges for offline planning, as simple discretization can effectively convert them to small discrete domains. However, even discretized domains like RS$(20,20)$ with only 3 observations remain extremely challenging due to their enormous state spaces.

Additionally, since these benchmarks are well-studied in the online planning community, we compare NVI's upper and lower bounds during backups with state-of-the-art online methods reported in the literature.
As shown in \Cref{fig:bounds_rs_ld}, NVI’s bounds gradually converge toward near-optimal solutions, consistent with the performance of online planners.
We omit the comparison for the Lidar Roomba problem, as we didn't find competitive performance in the current literature.

Last but not least, despite an extensive parameter search for Recurrent PPO (as detailed in the appendix), it failed to produce competitive policies compared to offline planning methods in most domains.
Notably, while some individual episodes during training showed promising behavior, the learning process frequently converged to a local optimum corresponding to a myopic solution (e.g., in RockSample, the agent moves directly to the exit area).
Notably, \citet{tao2025pobax} report higher performance in the RockSample problem, which may be attributed to differences in parameter settings or problem configurations.


\subsection{A Closer Look at Sampling-Based Approaches}

We examine the performance of NVI and POMCGS on the Light Dark domain, where both methods achieve similar results.
As shown in \Cref{fig:fsc_comparison,fig:time_comparison}, POMCGS is faster at finding reasonable solutions due to its top-down search and belief node merging.
However, this strategy has a key drawback: since it does not follow a value iteration scheme, its FSC nodes do not correspond to $\alpha$-vectors that generalize over large regions of the belief space.
As a result, the FSC can grow large and memory-heavy in complex domains.
In contrast, NVI performs value iteration and leverages the PWLC property, producing more compact policies. Moreover, NVI does not require state discretization as POMCGS does, since no belief comparison is needed.

\section{Discussion}

\paragraph{Contribution} In this article, we systematically demonstrate that a POMDP's value function can be represented with a finite set of neural networks. Based on this observation, we propose a new policy type, the finite network controller, a variation of the finite state controller in which each node stores a neural network to explicitly represent an $\alpha$-vector. We further propose a new offline POMDP planning algorithm called Neural Value Iteration, which follows the classic value iteration scheme but performs Bellman backups on neural networks. Through experiments, we demonstrate that this novel approach is able to solve one of the most challenging problems, RS($20,20$), which has more than 419 million states, far beyond the capability of existing offline solvers.

\paragraph{Limitation}
The main limitation of NVI is that the current approach only deals with continuous states. In continuous-state-action-observation POMDPs, compared with state-of-the-art offline methods like POMCGS, NVI also needs to discretize the observation space, same as POMCGS. However, NVI additionally needs to discretize the action space, while POMCGS uses action progressive widening.

\paragraph{Future Work}
The idea of representing $\alpha$-vectors with neural networks opens several directions for future research. Techniques such as state sampling \citep{pmlr-v28-brechtel13}, belief selection heuristics, macro actions \citep{macro_action_mcvi}, and observation grouping can be incorporated into Neural Value Iteration to further improve scalability in complex and continuous domains.

\section{Conclusion}

In the past decades, compared to the rapid advancement of learning approaches with neural networks, classic offline POMDP approaches have gradually received less attention due to scalability issues in complex domains.
However, in this article, we show that advancements in neural networks can also bring new life to the classic point-based value iteration scheme and may open a new direction for deep offline POMDP planning, highlighting that structured, model-based decision making still holds powerful potential.
In addition, many existing techniques from traditional value iteration can be naturally extended to Neural Value Iteration.
%

\bibliographystyle{plainnat}
\bibliography{refs-short,references}  

@article{lauri2022partially,
  title={Partially observable markov decision processes in robotics: A survey},
  author={Lauri, Mikko and Hsu, David and Pajarinen, Joni},
  journal={IEEE Transactions on Robotics},
  volume={39},
  number={1},
  pages={21--40},
  year={2022},
  publisher={IEEE}
}

@article{kurniawati2022partially,
  title={Partially observable markov decision processes and robotics},
  author={Kurniawati, Hanna},
  journal={Annual Review of Control, Robotics, and Autonomous Systems},
  volume={5},
  number={1},
  pages={253--277},
  year={2022},
  publisher={Annual Reviews}
}

@article{shani2013survey,
  title={A survey of point-based POMDP solvers},
  author={Shani, Guy and Pineau, Joelle and Kaplow, Robert},
  journal={Autonomous Agents and Multi-Agent Systems},
  volume={27},
  pages={1--51},
  year={2013},
  publisher={Springer}
}

@article{hansen1997improved,
  title={An improved policy iteration algorithm for partially observable MDPs},
  author={Hansen, Eric},
  journal={Advances in neural information processing systems},
  volume={10},
  year={1997}
}

@article{thrun1999monte,
  title={Monte carlo pomdps},
  author={Thrun, Sebastian},
  journal={Advances in neural information processing systems},
  volume={12},
  year={1999}
}

@article{spaan2005perseus,
  title={Perseus: Randomized point-based value iteration for POMDPs},
  author={Spaan, Matthijs TJ and Vlassis, Nikos},
  journal={Journal of artificial intelligence research},
  volume={24},
  pages={195--220},
  year={2005}
}

@article{porta2006point,
  title={Point-based value iteration for continuous POMDPs},
  author={Porta, Josep M and Vlassis, Nikos and Spaan, Matthijs TJ and Poupart, Pascal},
  journal={Journal of Machine Learning Research},
  volume={7},
  number={Nov},
  pages={2329--2367},
  year={2006}
}

@inproceedings{prentice2010belief,
  title={The belief roadmap: Efficient planning in linear POMDPs by factoring the covariance},
  author={Prentice, Sam and Roy, Nicholas},
  booktitle={Robotics Research: The 13th International Symposium ISRR},
  pages={293--305},
  year={2010},
  organization={Springer}
}

@inproceedings{bai2010monte,
  title={Monte Carlo value iteration for continuous-state POMDPs},
  author={Bai, Haoyu and Hsu, David and Lee, Wee Sun and Ngo, Vien A},
  booktitle={Algorithmic Foundations of Robotics IX: Selected Contributions of the Ninth International Workshop on the Algorithmic Foundations of Robotics},
  pages={175--191},
  year={2010},
  organization={Springer}
}

@article{moss2023betazero,
  title={BetaZero: Belief-state planning for long-horizon POMDPs using learned approximations},
  author={Moss, Robert J and Corso, Anthony and Caers, Jef and Kochenderfer, Mykel J},
  journal={arXiv preprint arXiv:2306.00249},
  year={2023}
}

@article{papadimitriou1987complexity,
  title={The Complexity of {Markov} Decision Processes},
  author={Papadimitriou, Christos H and Tsitsiklis, John N},
  journal={Mathematics of operations research},
  volume={12},
  number={3},
  pages={441--450},
  year={1987},
  publisher={INFORMS}
}

@inproceedings{madani1999undecidability,
  title={On the Undecidability of Probabilistic Planning and Infinite-Horizon Partially Observable {Markov} Decision Problems},
  author={Madani, Omid and Hanks, Steve and Condon, Anne},
  booktitle=aaai99,
  pages_={541--548},
  year={1999}
}

@InProceedings{Pineau-ijcai03,
  Author =	 {Pineau, Joelle and Gordon, Geoff and Thrun, Sebastian},
  Title =	 {Point-Based Value Iteration: An Anytime Algorithm for {POMDPs}},
  Year =	 2003,
  Publisher_ =	 {Morgan Kaufmann Publishers Inc.},
  BookTitle =	 ijcai03,
  Pages_ =	 {1025–1030},
  Location_ =	 {Acapulco, Mexico},
  Series_ =	 {IJCAI'03}
}

@InProceedings{Smith-uai04,
  Author =	 {Smith, Trey and Simmons, Reid},
  Title =	 {Heuristic Search Value Iteration for {POMDPs}},
  Year =	 2004,
  ISBN_ =	 0974903906,
  Publisher_ =	 {AUAI Press},
  BookTitle =	 uai04,
  Pages_ =	 {520–527},
  Location_ =	 {Banff, Canada},
}

@inproceedings{Smith_HSVI2,
author = {Smith, Trey and Simmons, Reid},
year = {2005},
month = {07},
pages = {},
title = {Point-Based {POMDP} Algorithms: Improved Analysis and Implementation},
booktitle = uai05
}

@inproceedings{sarsop,
  author =	 {Hanna Kurniawati and David Hsu and Wee Sun Lee},
  title =	 {{SARSOP}: Efficient point-based {POMDP} planning by approximating optimally reachable belief spaces},
  booktitle =	 rss08,
  year =	 2008
}

@inproceedings{macro_action_mcvi,
author = {Lim, Zhan Wei and Hsu, David and Lee, Wee Sun},
title = {Monte carlo value iteration with macro-actions},
year = {2011},
isbn = {9781618395993},
publisher = {Curran Associates Inc.},
address = {Red Hook, NY, USA},
booktitle = {Proceedings of the 25th International Conference on Neural Information Processing Systems},
pages = {1287–1295},
numpages = {9},
location = {Granada, Spain},
series = {NIPS'11}
}

@inproceedings{pomcp,
  author =	 {Silver, David and Veness, Joel},
  booktitle =	 nips10,
  publisher_ =	 {Curran Associates, Inc.},
  title =	 {{Monte-Carlo} Planning in Large {POMDPs}},
  volume =	 23,
  year =	 2010
}

@inproceedings{despot,
  author =	 {Somani, Adhiraj and Ye, Nan and Hsu, David and Lee, Wee Sun},
  booktitle =	 nips13,
  editor_ =	 {C.J. Burges and L. Bottou and M. Welling and Z. Ghahramani and K.Q. Weinberger},
  publisher_ =	 {Curran Associates, Inc.},
  title =	 {{DESPOT}: Online {POMDP} Planning with Regularization},
  volume_ =	 26,
  year =	 2013
}

@inproceedings{pomcpow,
author = {Sunberg, Zachary and Kochenderfer, Mykel},
year = {2017},
month = {09},
pages = {},
title = {{POMCPOW}: An online algorithm for {POMDP}s with continuous state, action, and observation spaces},
volume_ = {28},
booktitle = icaps17
}

@inproceedings{adaops,
 author = {Wu, Chenyang and Yang, Guoyu and Zhang, Zongzhang and Yu, Yang and Li, Dong and Liu, Wulong and Hao, Jianye},
 booktitle = nips21,
 editor_ = {M. Ranzato and A. Beygelzimer and Y. Dauphin and P.S. Liang and J. Wortman Vaughan},
 pages_ = {28419--28430},
 publisher_ = {Curran Associates, Inc.},
 title = {Adaptive Online Packing-guided Search for {POMDP}s},
 volume_ = {34},
 year = {2021}
}

@inproceedings{UCT,
author = {Kocsis, Levente and Szepesv\'{a}ri, Csaba},
title = {Bandit Based {Monte-Carlo} Planning},
year = {2006},
isbn_ = {354045375X},
publisher_ = {Springer-Verlag},
address_ = {Berlin, Heidelberg},
booktitle = ecml06,
pages_ = {282–293},
numpages_ = {12},
location_ = {Berlin, Germany},
series_ = {ECML'06}
}

@misc{drqn,
      title={Deep Recurrent Q-Learning for Partially Observable MDPs}, 
      author={Matthew Hausknecht and Peter Stone},
      year={2017},
      eprint={1507.06527},
      archivePrefix={arXiv},
      primaryClass={cs.LG},
      url={https://arxiv.org/abs/1507.06527}, 
}

@misc{ppo,
      title={Proximal Policy Optimization Algorithms}, 
      author={John Schulman and Filip Wolski and Prafulla Dhariwal and Alec Radford and Oleg Klimov},
      year={2017},
      eprint={1707.06347},
      archivePrefix={arXiv},
      primaryClass={cs.LG},
      url={https://arxiv.org/abs/1707.06347}, 
}

@misc{roomba,
  author = { K. Menda and Z. Sunberg and  M. Kochenderfer},
  title = {{RoombaPOMDPs.jl}},
  howpublished = "\url{https://github.com/sisl/RoombaPOMDPs.jl/}",
  year = {2023}, 
  note = "[Online; accessed 03-May-2025]"
}

@article{pomdpjl,
  author  = {Maxim Egorov and Zachary N. Sunberg and Edward Balaban and Tim A. Wheeler and Jayesh K. Gupta and Mykel J. Kochenderfer},
  title   = {POMDPs.jl: A Framework for Sequential Decision Making under Uncertainty},
  journal = jmlr,
  year    = {2017},
  volume  = {18},
  number  = {26},
  pages   = {1--5}
}

@inproceedings{you2025,
  title={{Partially Observable Monte-Carlo Graph Search}},
  author={You, Yang and Thomas, Vincent and Schutz, Alex and Skilton, Robert and Hawes, Nick and Buffet, Olivier},
  booktitle={Proceedings of the International Conference on Automated Planning and Scheduling},
  volume={35},
  pages={279--287},
  year={2025}
}

@article{karkus2017qmdp,
  title={Qmdp-net: Deep learning for planning under partial observability},
  author={Karkus, Peter and Hsu, David and Lee, Wee Sun},
  journal={Advances in neural information processing systems},
  volume={30},
  year={2017}
}

@InProceedings{pmlr-v28-brechtel13,
  title = 	 {Solving Continuous POMDPs: Value Iteration with Incremental Learning of an Efficient Space Representation},
  author = 	 {Brechtel, Sebastian and Gindele, Tobias and Dillmann, Rüdiger},
  booktitle = 	 {Proceedings of the 30th International Conference on Machine Learning},
  pages = 	 {370--378},
  year = 	 {2013},
  editor = 	 {Dasgupta, Sanjoy and McAllester, David},
  volume = 	 {28},
  series = 	 {Proceedings of Machine Learning Research},
  address = 	 {Atlanta, Georgia, USA},
  month = 	 {17--19 Jun},
  publisher =    {PMLR},
  url = 	 {https://proceedings.mlr.press/v28/brechtel13.html}
}

@article{tao2025pobax,
  author = {Tao, Ruo Yu and Guo, Kaicheng and Allen, Cameron and Konidaris, George},
  title = {Benchmarking Partial Observability in Reinforcement Learning with a Suite of Memory-Improvable Domains},
  booktitle = {Proceedings of the Second Reinforcement Learning Conference},
  journal = {The Reinforcement Learning Journal},
  url = {http://github.com/taodav/pobax},
  year = {2025},
}

@misc{jax2018github,
  author = {James Bradbury and Roy Frostig and Peter Hawkins and Matthew James Johnson and Chris Leary and Dougal Maclaurin and George Necula and Adam Paszke and Jake Vander{P}las and Skye Wanderman-{M}ilne and Qiao Zhang},
  title = {{JAX}: composable transformations of {P}ython+{N}um{P}y programs},
  url = {http://github.com/jax-ml/jax},
  version = {0.3.13},
  year = {2018},
}

@inproceedings{optuna_2019,
    title={Optuna: A Next-generation Hyperparameter Optimization Framework},
    author={Akiba, Takuya and Sano, Shotaro and Yanase, Toshihiko and Ohta, Takeru and Koyama, Masanori},
    booktitle={Proceedings of the 25th {ACM} {SIGKDD} International Conference on Knowledge Discovery and Data Mining},
    year={2019}
}

@book{sondik1971optimal,
  title={The optimal control of partially observable Markov processes},
  author={Sondik, Edward Jay},
  year={1971},
  publisher={Stanford University}
}

@article{cai2021hyp,
  title={HyP-DESPOT: A hybrid parallel algorithm for online planning under uncertainty},
  author={Cai, Panpan and Luo, Yuanfu and Hsu, David and Lee, Wee Sun},
  journal={The International Journal of Robotics Research},
  volume={40},
  number={2-3},
  pages={558--573},
  year={2021},
  publisher={SAGE Publications Sage UK: London, England}
}

@string{ijcai03 = "IJCAI-03"}

@string{ecml06 = "ECML-06"}

@string{icaps17 = "ICAPS-17"}

@string{aaai99 = "AAAI-99"}

@string{nips10 = "NIPS-10"}

@string{nips13 = "NIPS-13"}

@string{nips21 = "NIPS-21"}

@string{uai05 = "UAI-05"}

@string{uai04 = "UAI-04"}

@string{rss08 = "RSS-08"}

@string{nips = "NIPS"}

@string{jmlr = "JMLR"}

\section*{Appendix}

\subsection*{Belief Collection Process}

In our work, we adopt a standard belief collection technique similar to HSVI \cite{Smith_HSVI2} during value iteration.
The detailed procedure is provided in \Cref{alg:collect_beliefs}.
We maintain a belief tree $\mathcal{T}$ and perform forward search to collect beliefs along sampled trajectories.
At each step, actions are selected according to Q-values, and observations are chosen based on maximum excess uncertainty.
Upper bounds are initialized using a QMDP heuristic, while lower bounds are updated through value backups.

\begin{algorithm}[t]
\caption{\textbf{CollectBeliefs}}
\label{alg:collect_beliefs}
\SetKwInOut{Input}{Input}
\SetKwInOut{Output}{Output}

\Input{
  $\mathcal{T}$: the belief tree \\
  $n_{\mathcal{T}}$: current belief tree node \\
  $\epsilon$: uncertainty threshold \\
  $\gamma$: discount factor \\
  $d$: current depth \\
  $L$: maximum depth \\
  $nb_{\text{particle}}$: number of particles \\
  $V_{\text{mdp}}$: MDP heuristic \\
  $\mathcal{B}$: output list of sampled beliefs
}
\Output{$\mathcal{B}$: collected beliefs}

\BlankLine

\If{$\overline{V}(b_0) - \underline{V}(b_0) > \epsilon$ \textbf{and} $d < L$}{
    $a \gets \argmax_{a} Q_{\overline{V}}(n_{\mathcal{T}}, a)$ \;
    
    \ForEach{$o \in \Omega$}{
        $b' \gets \text{ParticleFilter}(nb_{\text{particle}}, n_{\mathcal{T}}.b, a, o)$ \;
        
        \If{$b' \notin \mathcal{T}$}{
            Add $b'$ to $\mathcal{T}$ as a child of $n_{\mathcal{T}}$ via edge $(a, o)$ \;
            Initialize $\overline{V}(b')$ using the QMDP heuristic with $V_{\text{mdp}}$ \;
        }
    }

    $o \gets \argmax_{o \in \Omega} ( \Pr(o|n_{\mathcal{T}}.b, a) \cdot \left( \overline{V}(b^{a,o}) - \underline{V}(b^{a,o}) - \epsilon \gamma^{-(d+1)} \right) )$ \;

    $\mathcal{B} \gets \mathcal{B} \cup \{b^{a,o}\}$ \;

    Recursively call $\text{CollectBeliefs}$ on the child node $n'_{\mathcal{T}}$ reached via edge $(a, o)$ from $n_{\mathcal{T}}$ with updated depth $d \gets d + 1$ \;
}

\Return $\mathcal{B}$
\end{algorithm}

\subsection*{RL Experiments}
For the RL experiments, we use the pobax package \citep{tao2025pobax}, which is entirely written in JAX \citep{jax2018github}\footnote{Commit 4e53aac is the latest version of the code we used}. RL experiments require significant amounts of fine-tuning hyperparameters, which is why JAX is currently the state-of-the-art choice, since it can run multiple runs in parallel on the GPU.

All experiments were performed on a Rocky Linux 9.6 (Blue Onyx) system with one NVIDIA A100-SXM4-40GB GPU and CUDA 12.9.

The network architecture of the actor critic model is the same as in \citet{tao2025pobax}, which is shown in Fig. \ref{fig:actor critic arcitecture}.
\begin{figure*}[t!]
    \centering
  \begin{lstlisting}[language=Python, basicstyle=\ttfamily\small]
ActorCritic(
  Actor(
    Sequential(
        (0): Dense(in_dims=input_dim, out_dims=hidden_size, bias=True)
        (1): ReLU()
        (3): GRU(in_dims=hidden_size, hidden_size=hidden_size)
        (4): Dense(in_dims=hidden_size, out_dims=hidden_size, bias=True)
        (5): ReLU()
        (6): Dense(in_dims=hidden_size, out_dims=action_dims, bias=True)
        (7): Categorical() or MultivariateNormalDiag()
    )
  )
  Critic(
    Sequential(
        (0): Dense(in_dims=input_dim, out_dims=hidden_size, bias=True)
        (1): ReLU()
        (3): GRU(in_dims=hidden_size, hidden_size=hidden_size)
        (4): Dense(in_dims=hidden_size, out_dims=hidden_size, bias=True)
        (5): ReLU()
        (6): Dense(in_dims=hidden_size, out_dims=1, bias=True)
    )
  )
)
\end{lstlisting}
    \caption{Actor critic architecture similar as \citeauthor{tao2025pobax}}
    \label{fig:actor critic arcitecture}
\end{figure*}
We performed hyperparameter tuning for each environment using the optuna package \citep{optuna_2019} with 50 trials and the following ranges:
\begin{itemize}
    \item Learning rate: $\text{lr} \in [10^{-5}, 10^{-3}]$ (log-uniform),
    \item Clipping parameter: $\epsilon \in [0.1, 0.4]$,
    \item Entropy coefficient: $\text{entropy\_coeff} \in [0.0, 0.1]$,
    \item Value function coefficient: $\text{vf\_coeff} \in [0.3, 0.7]$,
    \item GAE parameter: $\lambda_0 \in [0.1, 0.99]$,
    \item Gradient clipping: $\text{max\_grad\_norm} \in [0.5, 2.0]$,
    \item Parallel environments: $\text{num\_envs} \in \{64, 128, 256\}$,
    \item Rollout length: $\text{num\_steps} \in \{64, 128, 256\}$,
    \item Hidden layer size: $\text{hidden\_size} \in \{128, 256, 512, 1024\}$,
    \item PPO update epochs: $\text{update\_epochs} \in \{2, 3, \dots, 8\}$,
    \item Minibatches: $\text{num\_minibatches} \in \{2, 4, 8, 16\}$.
\end{itemize}

The best hyperparameters were chosen from the optuna results and are shown in table \ref{tab:nbvi_hyperparams}.

\begin{table*}[t!]
\centering
\caption{Best hyperparameters for PPO on various environments.}
\label{tab:nbvi_hyperparams}
\renewcommand{\arraystretch}{1.2}
\begin{tabular}{lcccccc}
\toprule
\textbf{Parameter} & {RS$(7, 8)$} &  {RS$(11, 11)$} &  {RS$(15, 15)$} & {RS$(20,20)$} & Light Dark  & Lidar Roomba\\
\midrule
Final Return              & 7.74      & 6.30     & 6.16     & 3.97     & 3.26        & -2.07 \\
\midrule
Learning Rate ($\alpha$) & 1.41e-4   & 5.08e-4  & 1.54e-4  & 2.42e-4  & 8.99e-4   & 3.81e-4 \\
Clip Epsilon ($\epsilon$)& 0.284     & 0.389    & 0.319    & 0.227    & 0.1024    & 0.3107 \\
Entropy Coeff. ($\beta$) & 0.084     & 0.015    & 0.076    & 0.0004   & 0.0126    & 0.0566 \\
Value Function Coeff. ($c_{vf}$) & 0.572 & 0.318 & 0.599    & 0.435    & 0.4793    & 0.4785 \\
GAE $\lambda_0$     & 0.367     & 0.448    & 0.102    & 0.734    & 0.1       & 0.1277 \\
Max Grad Norm            & 1.064     & 0.786    & 1.186    & 0.558    & 0.5627    & 1.9196 \\
Num Envs                 & 64        & 64       & 128      & 128      & 64        & 64 \\
Num Steps                & 64        & 64       & 64       & 64       & 64        & 256 \\
Hidden Size              & 256       & 128      & 256      & 128      & 512       & 256 \\
Update Epochs            & 3         & 6        & 8        & 6        & 7         & 6 \\
Num Minibatches          & 8         & 8        & 4        & 16       & 16        & 8 \\
\bottomrule
\end{tabular}
\end{table*}
For the RockSample environments, we used the implemented pobax implementations. Since there exists no JAX implementation of lightdark and roomba yet, we ported those from the Julia equivalents.
The full discounted reward during training is shown in Figure 2.

\begin{figure*}[t!]
    \centering
    \begin{subfigure}{0.32\linewidth}
        \includegraphics[width=\linewidth]{ 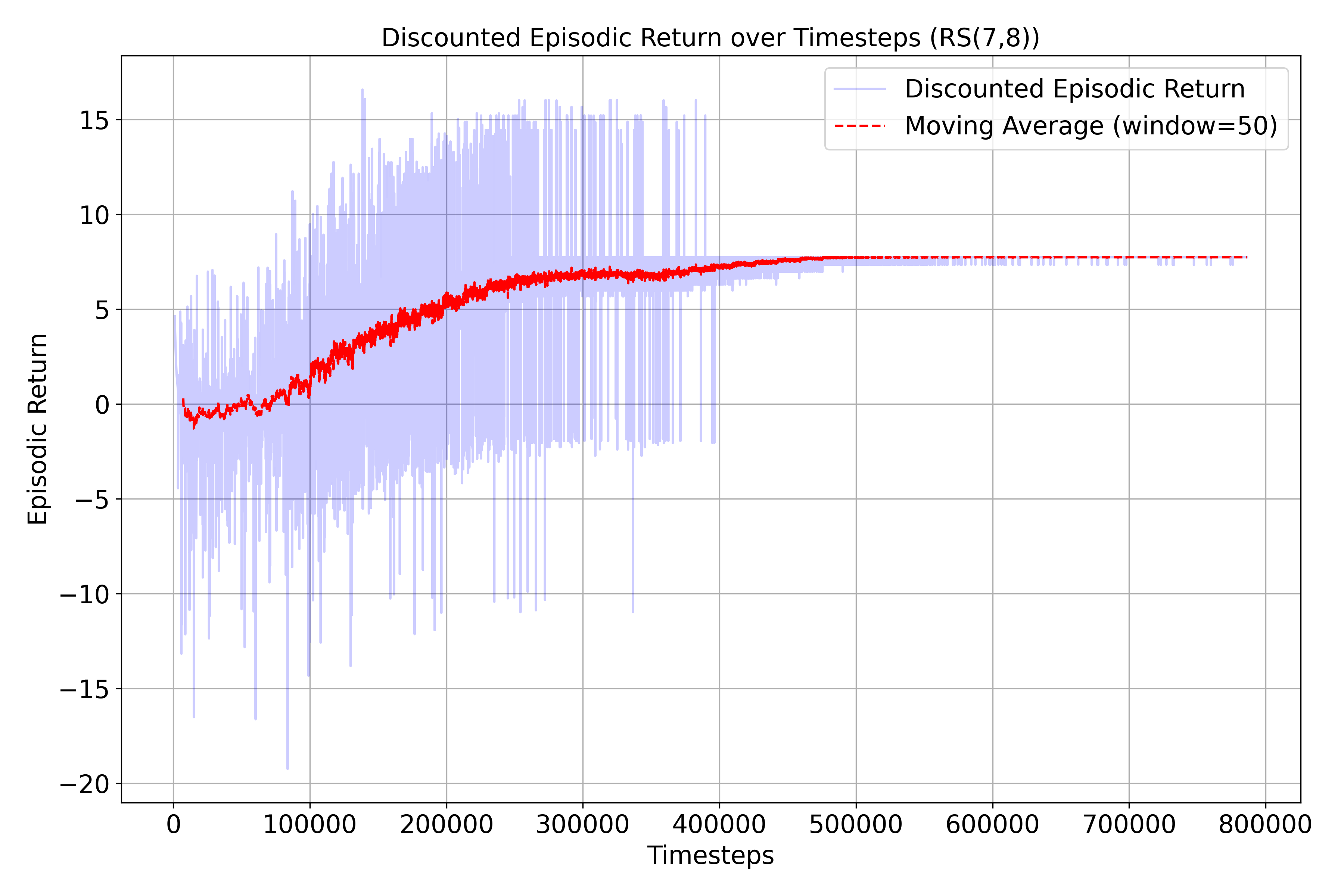}
        \caption{RS(7,8)}
        \label{fig:rs78}
    \end{subfigure}\hfill
    \begin{subfigure}{0.32\linewidth}
        \includegraphics[width=\linewidth]{ 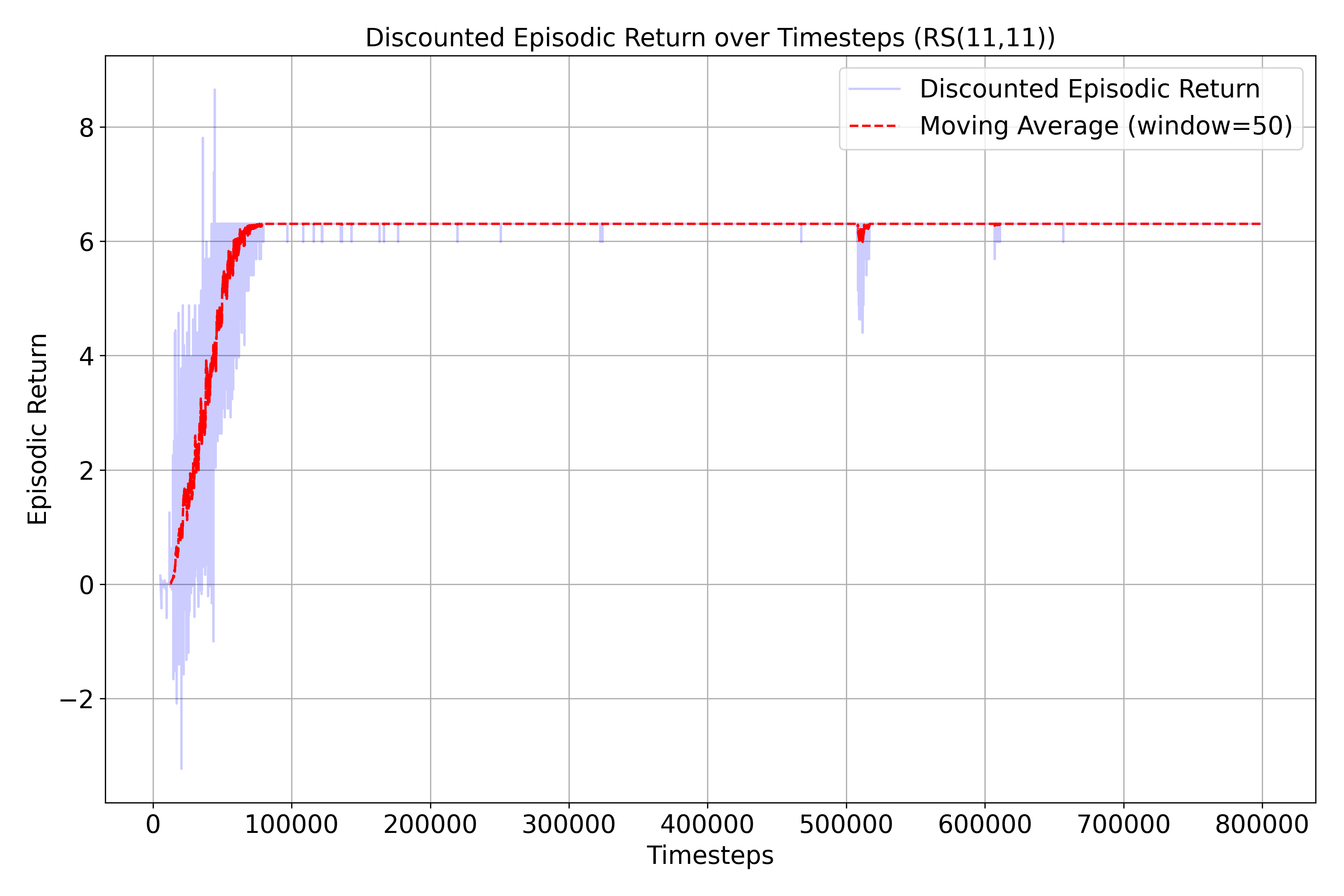}
        \caption{RS(11,11)}
        \label{fig:rs1111}
    \end{subfigure}\hfill
    \begin{subfigure}{0.32\linewidth}
        \includegraphics[width=\linewidth]{ 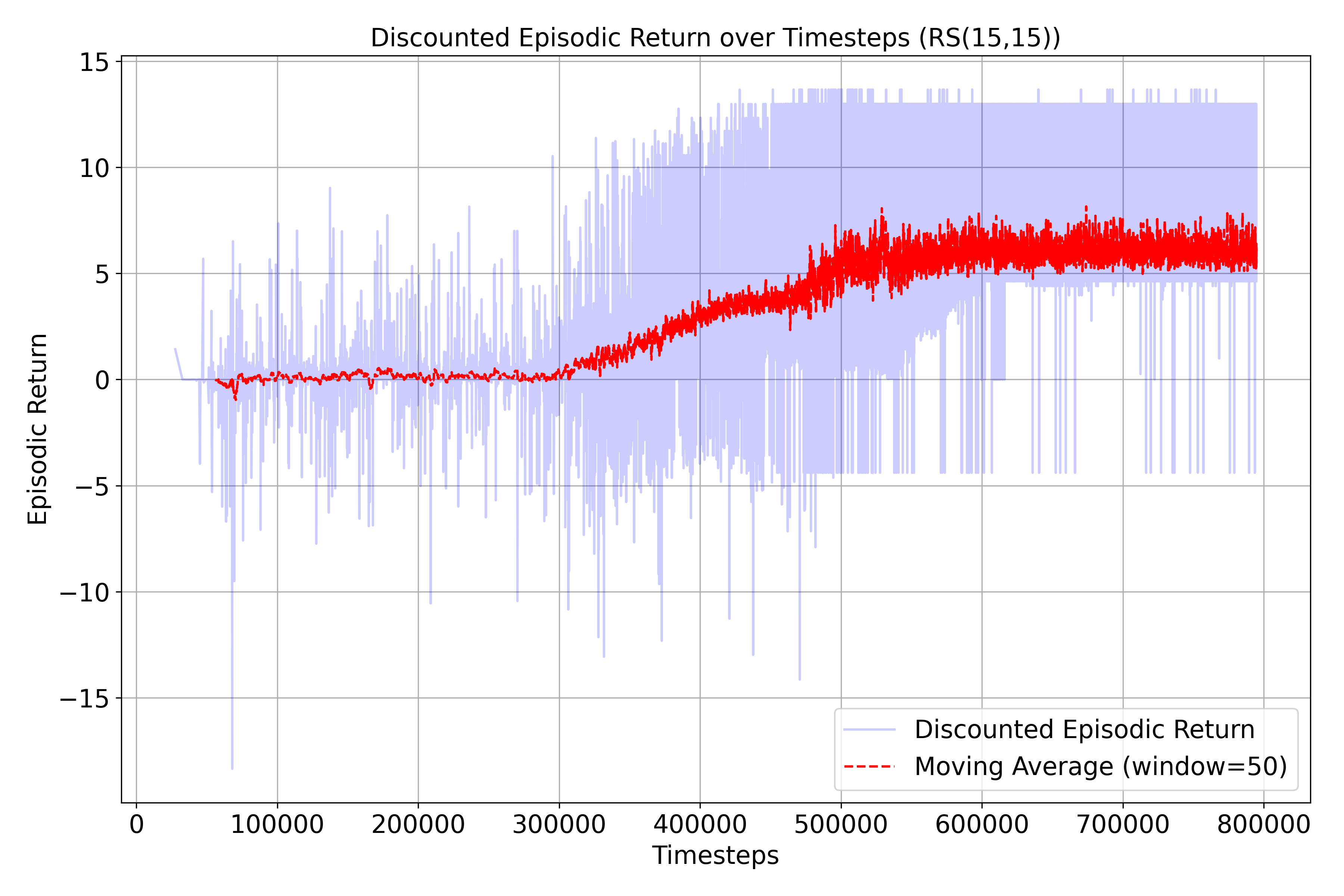}
        \caption{RS(15,15)}
        \label{fig:rs1515}
    \end{subfigure}

    \vspace{1em} 

    \begin{subfigure}{0.32\linewidth}
        \includegraphics[width=\linewidth]{ 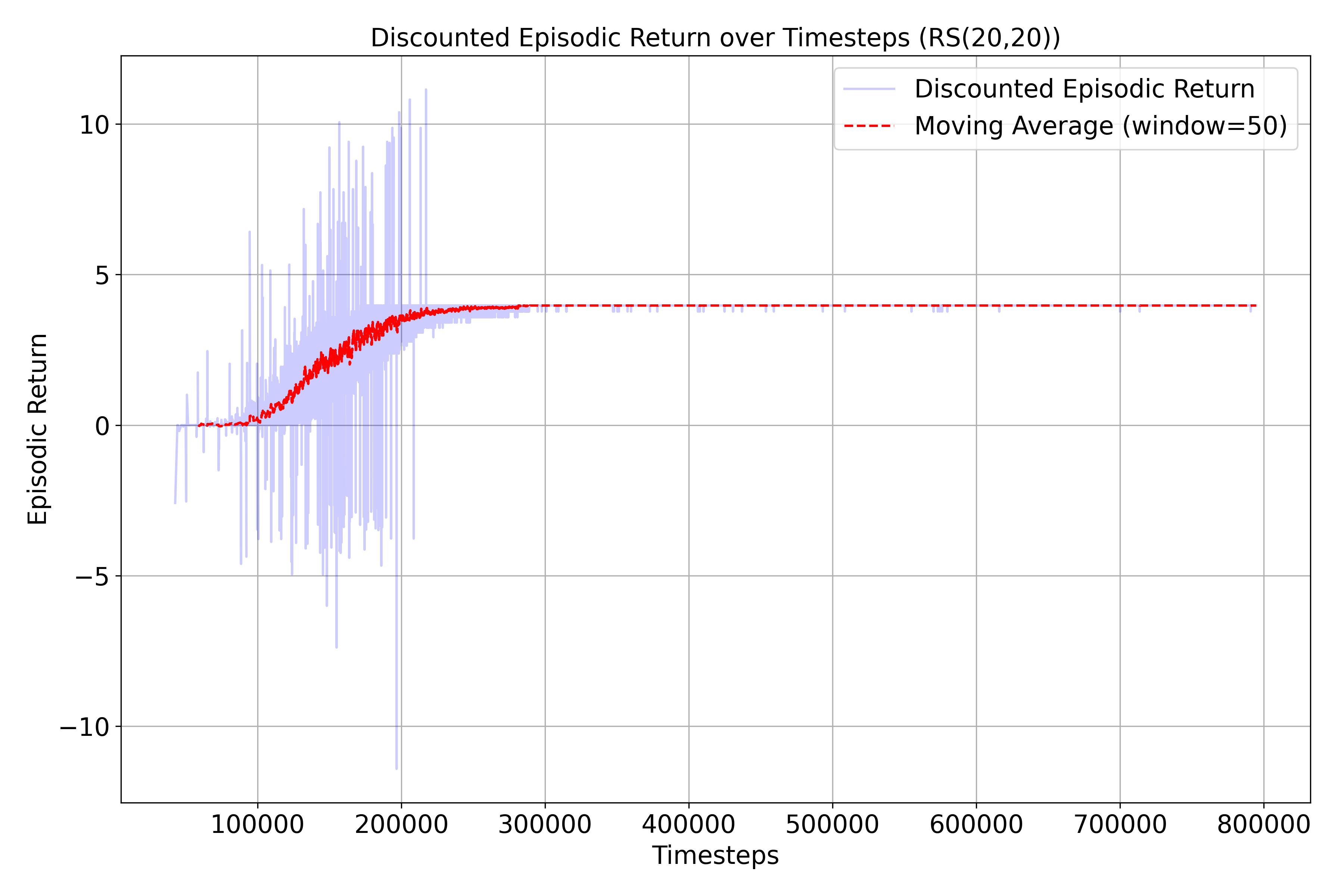}
        \caption{RS(20,20)}
        \label{fig:rs2020}
    \end{subfigure}\hfill
    \begin{subfigure}{0.32\linewidth}
        \includegraphics[width=\linewidth]{ 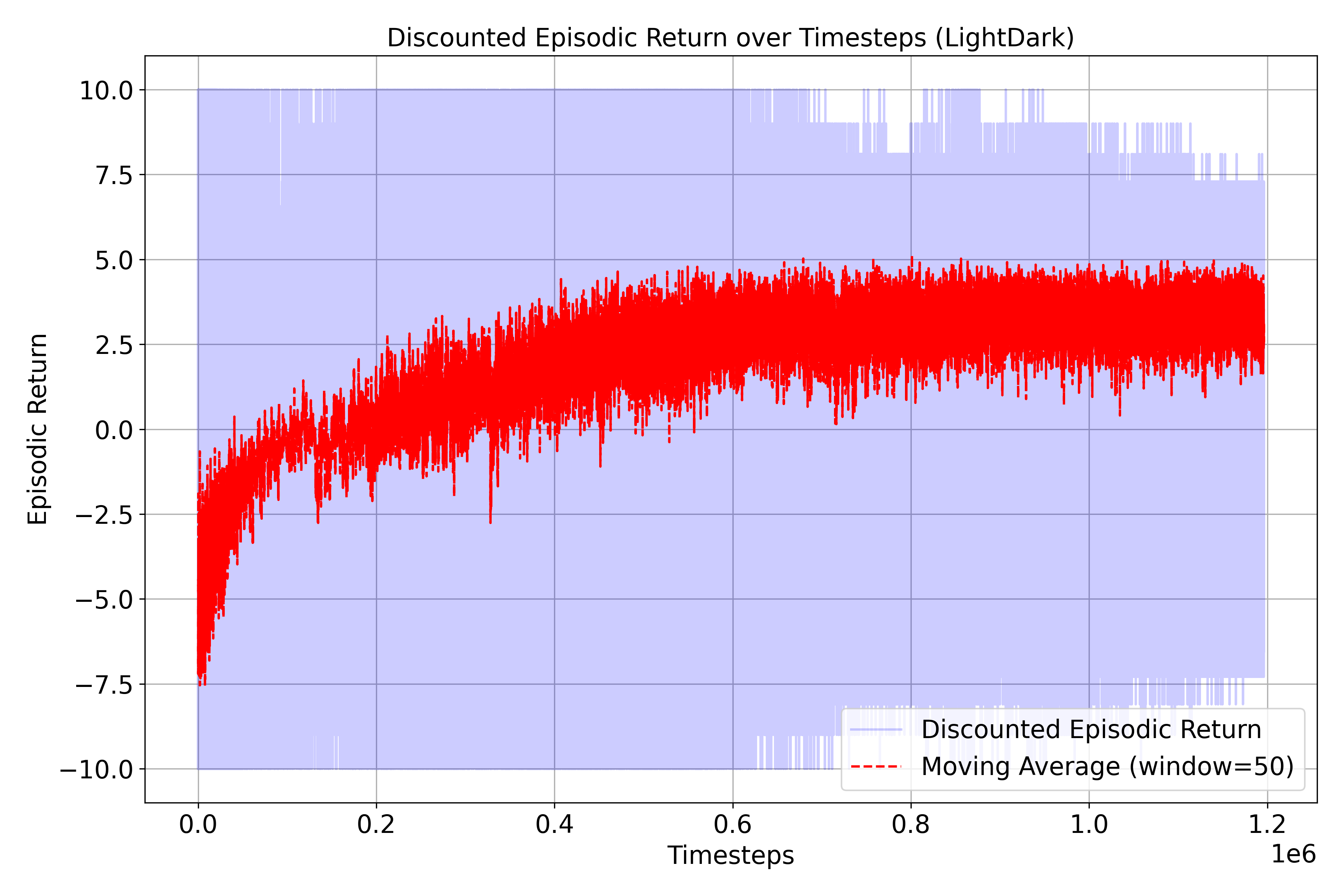}
        \caption{Light-Dark}
        \label{fig:lightdark}
    \end{subfigure}\hfill
    \begin{subfigure}{0.32\linewidth}
        \includegraphics[width=\linewidth]{ 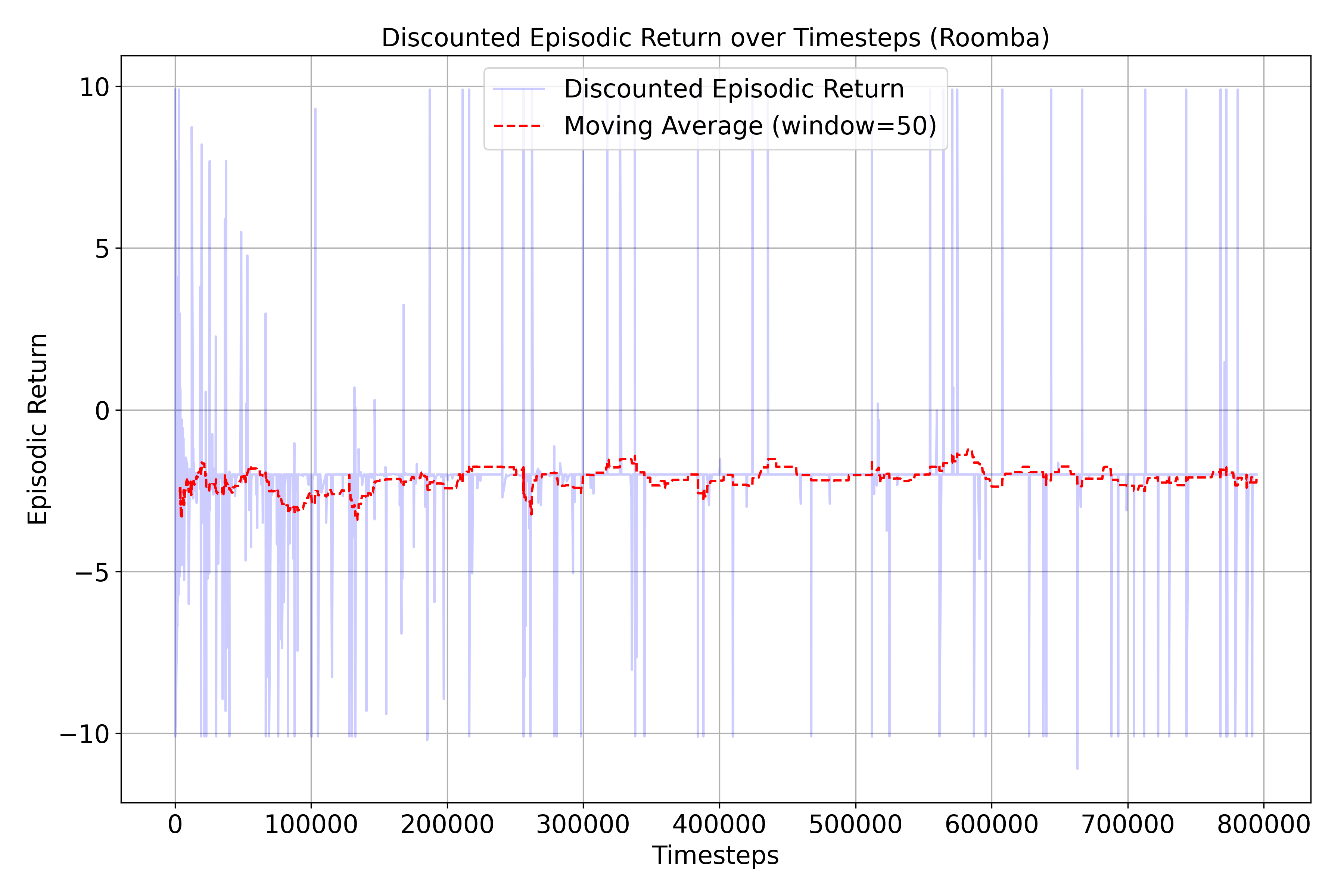}
        \caption{Lidar-Roomba}
        \label{fig:roomba}
    \end{subfigure}

    \caption{Training-progress curves for all evaluated environments.}
    \label{fig:training_progress_all}
\end{figure*}

\subsection*{Planning Experiments}
There are three planning algorithms compared in our experiments: SARSOP, POMCGS, and MCVI.
For POMCGS, we used the official POMCGS Julia implementation available at \url{https://github.com/ori-goals/POMCGraphSearch.jl}.
For MCVI, we implemented a customized version in Julia using the POMDP.jl framework.
We applied several custom optimizations, including vectorized operations, multi-threading, and static value types to accelerate computation.
These improvements enabled faster performance compared to existing Julia implementation (e.g., \url{https://github.com/JuliaPOMDP/MCVI.jl}) in our tested domains.

For SARSOP, we used the original C++ implementation available at \url{https://github.com/AdaCompNUS/sarsop}.
POMDP model files were exported in the \texttt{POMDPX} format for compatibility with SARSOP.

We used default settings for SARSOP and omit further parameter details.
For the sampling-based methods (POMCGS, MCVI, and NVI), all share two key parameters: $n_{\text{particle}}$, the number of particles used to represent beliefs, and $n_{\text{sim}}$, the number of simulations used for each state particle to approximate expectations.
We tested the following ranges:
\begin{itemize}
    \item $nb_{\text{particle}} \in \{1000, 5000, 10000, 20000, 50000\}$ 
    \item $nb_{\text{sim}} \in \{10, 50, 100, 500\}$
\end{itemize}
We found that $nb_{\text{particle}} = 10000$ and $nb_{\text{sim}} = 100$ were sufficient for good belief representation and expectation approximation across most benchmarks. These values were used in our final experiments.

For NVI, an additional parameter $nb_{\text{sample}}$ is used to sample the state space, and its impact is discussed in the main paper.

\paragraph{Network Architecture Selection} 
In NVI, the POMDP value function is approximated by a finite set of neural networks, each denoted as $\alpha_{\text{NN}}$.
In our experiments, we trained all $\alpha_{\text{NN}}$ with multilayer perceptrons (MLPs) with ReLU activations. 

We evaluated several candidate architectures with varying depth and width to balance approximation capacity and computational efficiency. 
The tested configurations included:

\begin{verbatim}
Dense(input_dim, 64, relu)
Dense(64, 64, relu)
Dense(64, 1)
\end{verbatim}

\begin{verbatim}
Dense(input_dim, 128, relu)
Dense(128, 64, relu)
Dense(64, 32, relu)
Dense(32, 1)
\end{verbatim}

\begin{verbatim}
Dense(input_dim, 128, relu)
Dense(128, 64, relu)
Dense(64, 64, relu)
Dense(64, 1)
\end{verbatim}

\begin{verbatim}
Dense(input_dim, 256, relu)
Dense(256, 128, relu)
Dense(128, 64, relu)
Dense(64, 1)
\end{verbatim}

\begin{verbatim}
Dense(input_dim, 512, relu)
Dense(512, 256, relu)
Dense(256, 256, relu)
Dense(128, 64, relu)
Dense(64, 1)
\end{verbatim}

Empirically, increasing network size beyond a moderate capacity did not produce noticeable improvements in policy quality. 
In contrast, smaller MLPs achieved comparable performance while reducing computational cost. 
In practice, the training time per backup was typically below 5 seconds, including CPU–GPU data transfer overhead with small MLPs.

Based on this empirical trade-off, we selected the following architectures:

\begin{itemize}
    \item \textbf{RockSample:}
    \begin{verbatim}
Dense(input_dim, 128, relu)
Dense(128, 64, relu)
Dense(64, 64, relu)
Dense(64, 1)
    \end{verbatim}

    \item \textbf{Light-Dark and Lidar Roomba:}
    \begin{verbatim}
Dense(input_dim, 128, relu)
Dense(128, 64, relu)
Dense(64, 32, relu)
Dense(32, 1)
    \end{verbatim}
\end{itemize}
We fixed the learning rate to 0.005 and used a batch size of 1024 for training each $\alpha_{\text{NN}}$.
Training stops early if the mean squared error (MSE) falls below 0.1 or after a maximum of 10000 epochs.

\paragraph{Input and Output of Network}
For each $\alpha_{\text{NN}}$, the input is a vectorized state representation and the output is a scalar, consistent with the definition of an $\alpha$-vector as a mapping from $\mathbb{S}$ to $\mathbb{R}$. Specifically, in RockSample, the input consists of a one-hot encoding of the robot position concatenated with the rock status vector. In LightDark and LidarRoomba, the input is the state vector representing the agent’s location. 

\end{document}